\documentclass[10pt,twocolumn,letterpaper]{article}

\usepackage{cvpr}

\newcommand{\PAR}[1]{\vspace{0.1cm}\noindent{\bf #1} }

\definecolor{tabfirst}{rgb}{1, 0.7, 0.7} %
\definecolor{tabsecond}{rgb}{1, 0.85, 0.7} %
\definecolor{tabthird}{rgb}{1, 1, 0.7} %

\definecolor{cvprblue}{rgb}{0.21,0.49,0.74}
\usepackage[pagebackref,breaklinks,colorlinks,allcolors=cvprblue]{hyperref}

\usepackage[capitalize]{cleveref}
\crefname{section}{Sec.}{Secs.}
\Crefname{section}{Section}{Sections}
\Crefname{table}{Table}{Tables}
\crefname{table}{Tab.}{Tabs.}

\usepackage{algorithm}
\usepackage{algorithmic}
\usepackage{caption}
\usepackage{multicol}
\usepackage{multirow}
\usepackage{marvosym}

\title{Convex Relaxation for Robust Vanishing Point Estimation in Manhattan World}

\author{Bangyan Liao$^{1,2\star}$, Zhenjun Zhao$^{3\star}$, Haoang Li$^{4}$, Yi Zhou$^{5}$, Yingping Zeng$^{5}$, Hao Li$^{1}$, Peidong Liu$^{1}\textsuperscript{\Letter}$\\
	\\
	$^{1}$Westlake University\quad$^{2}$Zhejiang University\quad
	$^{3}$University of Zaragoza\\
	$^{4}$Hong Kong University of Science and Technology (Guangzhou) \quad
	$^{5}$Hunan University
}

\begin{document}

\maketitle

\let\thefootnote\relax\footnotetext{$^\star$ Equal contribution.} 
\footnotetext{$^\text{\Letter}$ Corresponding author: Peidong Liu (liupeidong@westlake.edu.cn).}

\begin{abstract}
Determining the vanishing points (VPs) in a Manhattan world, as a fundamental task in many 3D vision applications, consists of jointly inferring the line-VP association and locating each VP. Existing methods are, however, either sub-optimal solvers or pursuing global optimality at a significant cost of computing time. In contrast to prior works, we introduce convex relaxation techniques to solve this task for the first time. Specifically, we employ a ``soft'' association scheme, realized via a truncated multi-selection error, that allows for joint estimation of VPs' locations and line-VP associations. This approach leads to a primal problem that can be reformulated into a quadratically constrained quadratic programming (QCQP) problem, which is then relaxed into a convex semidefinite programming (SDP) problem. To solve this SDP problem efficiently, we present a globally optimal outlier-robust iterative solver (called \textbf{GlobustVP}), which independently searches for one VP and its associated lines in each iteration, treating other lines as outliers. After each independent update of all VPs, the mutual orthogonality between the three VPs in a Manhattan world is reinforced via local refinement. Extensive experiments on both synthetic and real-world data demonstrate that \textbf{GlobustVP} achieves a favorable balance between efficiency, robustness, and global optimality compared to previous works. The code is publicly available at \href{https://github.com/WU-CVGL/GlobustVP/}{github.com/wu-cvgl/GlobustVP}.
\end{abstract} 
\section{Introduction}
\label{sec:intro}

\begin{figure}[h]
	\centering
	\begin{subfigure}[b]{0.48\linewidth}
		\centering
        \caption*{\centering Ground Truth}
		\includegraphics[width=\textwidth]{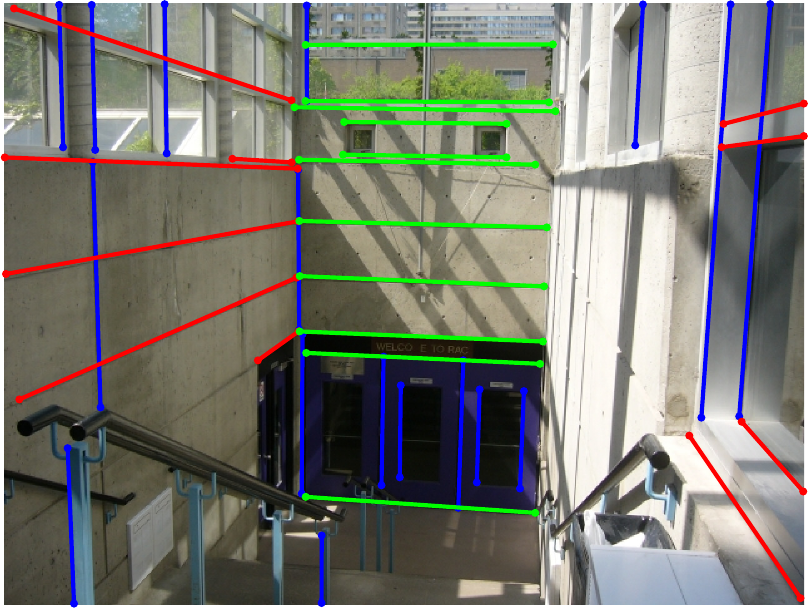}
		\caption*{\centering 3 VPs \par 38 lines}
		\label{fig_teaser_gt}
	\end{subfigure}
	\begin{subfigure}[b]{0.48\linewidth}
		\centering
        \caption*{\centering \textbf{GlobustVP (Ours)}}
		\includegraphics[width=\textwidth]{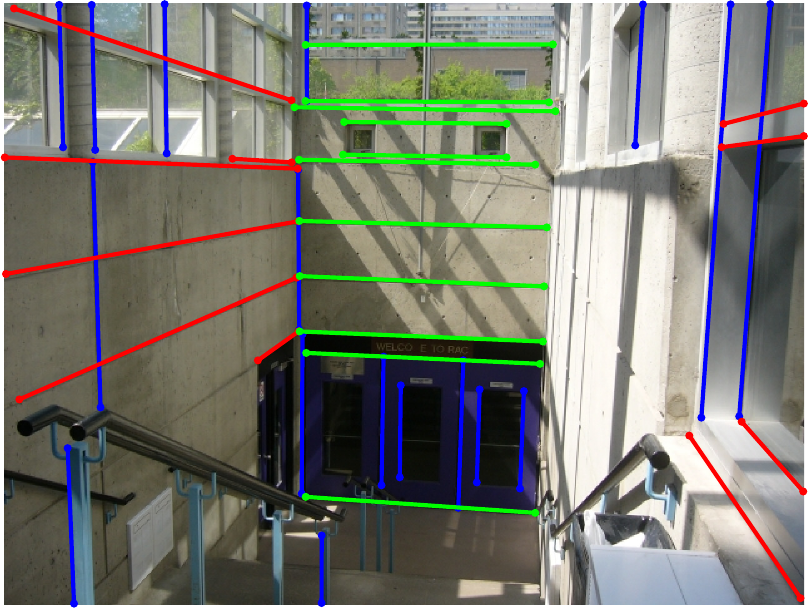}
        \caption*{\centering \textcolor{teal}{\textbf{$100\%$}}, \textcolor{purple}{\textbf{$100\%$}} \par \textcolor{violet}{\textbf{0.23 pix.}}, \textcolor{orange}{\textbf{50 ms}}}
		\label{fig_teaser_ours}
	\end{subfigure}
    \caption{Representative results of our method ({\it \textbf{GlobustVP}}) on the York Urban Database~\cite{Denis2008EfficientEM} for estimating three vanishing points (VPs) in the Manhattan world. Different line-VP associations are shown in respective colors. Below each image, we report the widely-used evaluation metrics, \textcolor{teal}{precision $\uparrow$}, \textcolor{purple}{recall $\uparrow$}, \textcolor{violet}{consistency error $\downarrow$} (defined in~\cref{sec:Experiments:Evaluation metrics}), and \textcolor{orange}{runtime $\downarrow$}. {\it \textbf{GlobustVP}} achieves high precision and fast runtime, demonstrating an excellent balance between accuracy and efficiency.}
	\label{fig_teaser}
\end{figure}

Multiple parallel 3D lines project onto the camera image plane, converging at a specific point known as the vanishing point (VP). Vanishing point estimation is a fundamental task in Simultaneous Localization and Mapping (SLAM)~\cite{Li2009ConsensusSM,li2019leveraging}, camera calibration~\cite{lee2013automatic, li2019line}, and structural understanding~\cite{flint2011manhattan}. In this paper, we focus on vanishing point estimation in the Manhattan world~\cite{straub2017manhattan}, which assumes only three mutually orthogonal line directions and represents the most common structural pattern in man-made environments. The challenge lies in the coupling relationship between line association and vanishing point estimation, which leads to numerous local minima. This characteristic makes an algorithm with a global optimality guarantee particularly valuable.

Previous methods for accurate and efficient vanishing point estimation rely on local, global, or learning-based approaches. Local methods (\eg, Expectation-Maximization~\cite{Denis2008EfficientEM}, Hough voting~\cite{Wu2021RealtimeVP}, and RANSAC~\cite{Fischler1981RandomSC}) require either a good VP initialization or exhaustive sampling to achieve the desired accuracy. Due to the stochastic nature of local solvers, they cannot guarantee global optimality. In contrast, global methods (\eg, Branch-and-Bound~\cite{Bazin2012GloballyOL,Li2019QuasiGloballyOA}) always guarantee the globally optimal solution. However, the exponential time cost in the worst case makes them inefficient, and the large search space hinders their use in large-scale scenarios. Recently, the success of deep learning has led researchers to apply it to vanishing point estimation~\cite{lin2022deep, tong2022transformer}. However, learning-based methods lack optimality guarantees and often struggle to generalize beyond their training datasets, limiting their practical utility. In conclusion, a globally optimal, outlier-robust, and sufficiently fast vanishing point estimation algorithm remains missing.

To address this, we introduce convex relaxation techniques for the first time in the context of vanishing point estimation. Specifically, we propose a novel truncated multi-selection error formulation that enables joint optimization of three vanishing points and their corresponding line association labels (\eg, VP1, VP2, VP3, Outlier). This error formulation is incorporated into~\ref{eq:Primal Problem}, reformulated as~\ref{eq:Full QCQP Problem}, and subsequently relaxed into the convex \ref{eq:Full SDP Problem}. To solve such a problem efficiently, we propose an iterative solver {\it \textbf{GlobustVP}}, which solves one~\ref{eq:Single Block SDP Problem} in each iteration to identify a vanishing point and its corresponding inlier line set, which is then excluded from subsequent iterations. After finding all three vanishing points, \ref{eq:Manhattan Post-Refinement} is applied to ensure that the estimated vanishing points satisfy the Manhattan world assumption.

In summary, our \textbf{contributions} are as follows:
\begin{itemize}
    \item A ``soft'' association scheme, realized via the proposed truncated multi-selection error, for joint estimation of VPs' location and line-VP association.
    \item Introducing convex relaxation to reformulate the intermediate QCQP form of the primal problem as a convex SDP problem. 
    \item An iterative solver, {\it \textbf{GlobustVP}}, that solves each VP sub-problem independently (corresponding to a sub-block of the full SDP problem), achieving global optimality under mild conditions.
    \item Extensive evaluations on both synthetic and real data demonstrate that our method achieves superior accuracy and robustness, while being on par with prior methods in terms of efficiency.
\end{itemize}
\section{Related Work}
\label{sec:Related Work}
Existing vanishing point estimation methods can be broadly categorized into five main classes: Expectation-Maximization (EM)-based~\cite{Antone2000AutomaticRO,Denis2008EfficientEM,Coughlan2003ManhattanWO}, Hough-voting-based~\cite{barnard1982interpreting,Quan1989DeterminingPS,Gamba1996VanishingPD,Wu2021RealtimeVP,Matessi1999VanishingPD,Lutton1994ContributionTT}, RANSAC-based~\cite{Zuliani2005TheMA,Sinha2008Interactive3A,Toldo2008RobustMS,Bazin20123lineRF,Mirzaei2011OptimalEO,Zhang2016VanishingPE,Tardif2009NoniterativeAF}, Branch-and-Bound (BnB)-based~\cite{Bazin2012GloballyOL,Bazin2012GloballyOC,Li2020QuasiGloballyOA,Li2019QuasiGloballyOA}, and learning-based methods~\cite{zhai2016detecting,Zhou2019NeurVPSNV,Kluger2020CONSACRM,li2021learning,lin2022deep,tong2022transformer}.

EM-based methods~\cite{Antone2000AutomaticRO,Denis2008EfficientEM,Coughlan2003ManhattanWO} iteratively associate image lines to estimate vanishing points, but they are prone to local minima and sensitive to initialization.
Hough-voting-based approaches~\cite{barnard1982interpreting,Quan1989DeterminingPS,Gamba1996VanishingPD} partition the Gaussian sphere and accumulate votes along the locus of each great circle to identify the dominant directions. However, they are susceptible to bin quantization issues and may detect spurious vanishing points.
RANSAC-based methods~\cite{Toldo2008RobustMS,Zuliani2005TheMA} typically sample subsets of image lines to hypothesize candidate vanishing points and evaluate each hypothesis based on inlier consensus. While achieving robustness against outliers, these approaches are inherently non-deterministic and do not guarantee global optimality due to sampling uncertainty and failure to satisfy orthogonality constraints~\cite{Tardif2009NoniterativeAF,Mirzaei2011OptimalEO}.
BnB-based approaches~\cite{Bazin2012GloballyOL,Li2019QuasiGloballyOA,Li2020QuasiGloballyOA} treat vanishing point estimation as a global optimization problem and use strategies such as branch-and-bound~\cite{li2009consensus,hartley2009global} to improve performance. Bazin~\etal~\cite{Bazin2012GloballyOL} formulate the task as a consensus set maximization problem over the rotation search space and further solve it efficiently using a BnB scheme. While these methods guarantee global optimality, they can be computationally expensive due to the exploratory nature and exponential worst-case performance of BnB. Hybrid approaches~\cite{Li2019QuasiGloballyOA,Li2020QuasiGloballyOA} combine BnB and RANSAC to improve efficiency, but only achieve quasi-global optimality.

Recently, learning-based methods have demonstrated effectiveness in vanishing point estimation. Zhai~\etal~\cite{zhai2016detecting} use a deep convolutional neural network (CNN) to extract the global image context and generate horizon line candidates. NeurVPS~\cite{Zhou2019NeurVPSNV} employs conic convolutions to refine feature extraction along structural lines. Lin~\etal~\cite{lin2022deep} incorporate the Hough Transform and Gaussian sphere into a deep vanishing point detection network. More recently, methods like CONSAC~\cite{Kluger2020CONSACRM} and PARSAC~\cite{kluger2024parsac} combine neural networks with multi-model fitting to improve vanishing point estimation. While these methods achieve promising results, they often involve trade-offs between computational efficiency and generalization, particularly when compared to traditional RANSAC- or BnB-based approaches.

In summary, achieving a balance between computational efficiency, robustness to outliers, and global optimality remains a key challenge in vanishing point estimation. Our approach addresses these limitations by incorporating a convex relaxation technique, offering a favorable trade-off between efficiency, robustness, and optimality. Extensive experiments further demonstrate the effectiveness of our method across a variety of settings.
\section{Notations and Background}
\label{sec:Background Theory}
\subsection{Notation}
\label{subsec:Notation}

We use lowercase letters (\eg, $s$) to denote scalars, bold lowercase letters (\eg, $\mathbf{v}$) for vectors, and bold uppercase letters (\eg, $\mathbf{M}$) for matrices. $[\mathbf{M}]_{i,j}$ denotes the scalar entry in the $i^{\text{th}}$ row and $j^{\text{th}}$ column of the matrix $\mathbf{M} \in \mathbb{R}^{m \times n}$. $\{\mathbf{M}\}_{i,j}$ denotes the block sub-matrix at the $i^{\text{th}}$ row and $j^{\text{th}}$ column of the matrix $\mathbf{M}$. $[\mathbf{M}]_{i,*}$ denotes the $i^{\text{th}}$ row of the matrix $\mathbf{M}$. The operation $(\mathbf{M})^2: \mathbb{R}^{m \times n} \rightarrow \mathbb{R}^{m \times n}$ denotes the element-wise square operator. $\mathbf{I}_d$ is the identity matrix of size $d$. The vectors ${\mathbf{1}}_{n} \in \mathbb{R}^n$ and ${\mathbf{0}}_{n} \in \mathbb{R}^n$ denote all-one and all-zero vectors, respectively. The $[\mathbf{M} ;\mathbf{N}]$ operator denotes the matrix stacking operator, while the $[\mathbf{M}, \mathbf{N}]$ operator denotes matrix concatenation. We use ``$\otimes$'' to denote the Kronecker product. $\text{vec}(\mathbf{M})$ denotes the vectorization operator, which concatenates the columns of $\mathbf{M}$. $\text{diag}([\mathbf{M}_1,\ldots, \mathbf{M}_n])$ denotes the operation of arranging the square matrices $\mathbf{M}_1$ to $\mathbf{M}_n$ along the diagonal.

\subsection{Geometry Related to Vanishing Point}
\label{subsec:Geometry Related to Vanishing Point}

The vanishing point is a widely used concept in 3D vision, based on the observation that two parallel lines in a Euclidean space intersect at a point in the image projection space, known as the vanishing point~\cite{hartley2003multiple,Zhang2016VanishingPE}. This constraint can be expressed as: 
\begin{align}
\mathbf{v}^{\top}\mathbf{l} = 0,
\end{align}
where $\mathbf{v} \in \mathbb{R}^3$ and $\mathbf{l} \in \mathbb{R}^3$ denote the homogeneous vanishing point and its associated line on image plane, respectively. We can convert $\mathbf{v}$ and $\mathbf{l}$ to the normalized image coordinate system:
\begin{align}
\mathbf{d}^{\top}\mathbf{n} &= 0,\\
\mathbf{v}=\mathbf{K}\mathbf{d}, \quad&\quad \mathbf{l}=\mathbf{K}^{-\top}\mathbf{n},
\end{align}
where $\mathbf{K} \in \mathbb{R}^{3 \times 3}$ is the camera intrinsic matrix~\cite{hartley2003multiple},  $\mathbf{d} \in \mathbb{R}^3$ represents the 3D line direction (equivalent to the vanishing point), and $\mathbf{n} \in \mathbb{R}^3$ is the normal vector to the projection plane for each image line. For simplicity, we will adopt the normalized coordinate system in the following derivations.
Additionally, the Manhattan world assumption implies that the directions of the three 3D lines $\mathbf{d}$ are mutually orthogonal, forming a $3 \times 3$ orthogonal matrix. Leveraging this Manhattan world prior constraint can enhance the stability of the overall optimization algorithm.

\subsection{Convex Relaxation}
\label{sec:Convex Relaxation}
Given a general Quadratically Constrained Quadratic Programming (QCQP) formulation:
\begin{equation}
    \begin{aligned}
        \min_{\mathbf{x} \in \mathbb{R}^n} \quad & \mathbf{x}^{\top} \mathbf{C} \mathbf{x}\\
         \text{s.t. } \quad & \mathbf{x}^{\top} \mathbf{A}_i \mathbf{x} = b_i, \quad i = 1, \ldots, m,
    \end{aligned}
\end{equation}
where $\mathbf{C}, \mathbf{A}_1, \dots,\mathbf{A}_m \in \mathbb{S}^n$ are real symmetric matrices. Using the matrix trace property, we can convert the quadratic error term as:
\begin{align}
\textbf{x}^{\top} \textbf{C} \textbf{x} = \text{trace}(\textbf{x}^{\top} \textbf{C} \textbf{x}) = \text{trace}(\textbf{C}\textbf{x}  \textbf{x}^{\top}).
\end{align}

Thus, by introducing a new rank-one symmetric positive semidefinite (PSD) matrix $\mathbf{X} = \mathbf{x} \mathbf{x}^{\top}$, we can reformulate the above QCQP as:
\begin{equation}
    \begin{aligned}
        \min_{\mathbf{X} \in \mathcal{S}^n} \quad & \text{trace}(\mathbf{C}\mathbf{X}) \\
         \text{s.t. } \quad & \text{trace}(\mathbf{A}_i\mathbf{X}) = b_i, \quad i = 1, \ldots, m, \\
         & \mathbf{X} \succeq \mathbf{0}, \quad \text{rank}(\mathbf{X}) = 1.
    \end{aligned}
\end{equation}

By relaxing the rank constraint, we obtain the standard semidefinite programming (SDP) formulation:
\begin{equation}
    \begin{aligned}
        \min_{\mathbf{X} \in \mathcal{S}^n} \quad & \text{trace}(\mathbf{C}\mathbf{X})\\
         \text{s.t. } \quad & \text{trace}(\mathbf{A}_i\mathbf{X}) = b_i, \quad i = 1, \ldots, m, \\
         & \mathbf{X} \succeq \mathbf{0}.
            \end{aligned}
\end{equation}

The primary benefit of employing the convex relaxation, rather than solving the original problem, lies in the convex nature of the resulting formulation. Although relaxation has been applied, researchers have found that strong duality properties still hold for many problems~\cite{briales2017convex, zhao2020efficient}. This implies that solving the relaxed problem is equivalent to solving the original one~\cite{anstreicher2012convex}. Leveraging any off-the-shelf SDP solver~\cite{MOSEK}, we can always find the global optimum in polynomial time.
\section{Methodology}
\label{sec:Methodology}

In~\cref{subsec:Truncated Multi-selection Error Formulation}, we formulate the vanishing point estimation and line association problem as~\ref{eq:Primal Problem} using the proposed truncated multi-selection error. In~\cref{subsec:Full QCQP Formulation and Convex Relaxation}, we reformulate this~\ref{eq:Primal Problem} as~\ref{eq:Full QCQP Problem} and subsequently rewrite it as \ref{eq:Full SDP Problem}. To solve such a SDP problem efficiently, we introduce an iterative solver {\it \textbf{GlobustVP}} in ~\cref{subsec:GlobustVP}, which solves one instance of \ref{eq:Single Block SDP Problem} per iteration, followed by \ref{eq:Manhattan Post-Refinement} after finding all three VPs.

\subsection{Truncated Multi-selection Error Formulation}
\label{subsec:Truncated Multi-selection Error Formulation}

\begin{figure}[tb]
  \centering
    \begin{subfigure}[b]{0.48\linewidth}
		\centering
		\includegraphics[width=\linewidth]{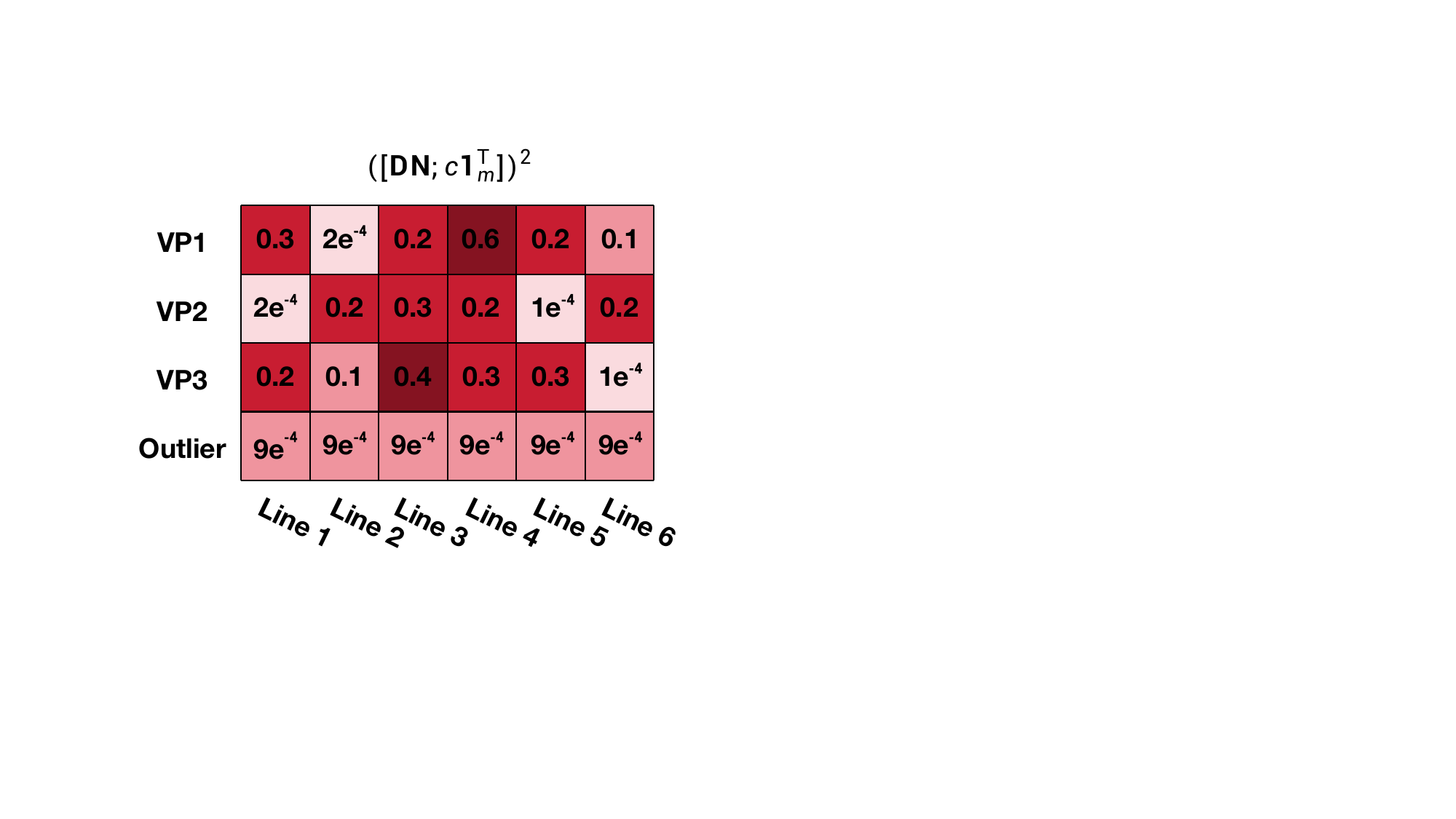}
		\caption{} \label{fig:DN_example}
    \end{subfigure}
    \begin{subfigure}[b]{0.48\linewidth}
		\centering
		\includegraphics[width=\linewidth]{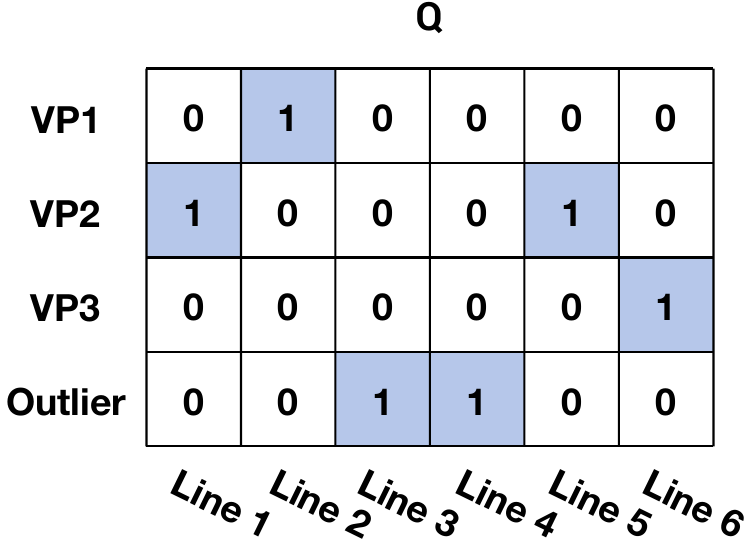}
		\caption{} \label{fig:Q_example}
    \end{subfigure}
  \caption{An example of the distance matrix $([\mathbf{DN}; c\mathbf{1}_m^{\top}])^2$ and the associated permutation matrix $\mathbf{Q}$. In this example, $c=0.03$, line 1 matches VP2, line 2 matches VP1, line 3 and 4 are identified as outliers, line 5 matches VP2, and line 6 matches VP3.}
  \label{fig:Q_DN_example}
\end{figure}

We begin by defining the VP matrix $\mathbf{D} = [\mathbf{d}_1^{\top}; \mathbf{d}_2^{\top}; \mathbf{d}_3^{\top}] \in \mathbb{R}^{3 \times 3}$, where each $\mathbf{d}_i \in \mathbb{R}^3$ represents a 3D line direction (equivalent to the vanishing point). Next, we define the line matrix $\mathbf{N} = [\mathbf{n}_1, \mathbf{n}_2, ..., \mathbf{n}_m] \in \mathbb{R}^{3 \times m}$, where each $\mathbf{n}_i \in \mathbb{R}^3$ represents a unit normal perpendicular to the projection plane (equivalent to an image line), and $m$ denotes the number of lines in the image. By multiplying the VP matrix $\mathbf{D}$ with the line matrix $\mathbf{N}$, stacking it with $c \mathbf{1}_{m}^\top$, and then performing element-wise squaring, we obtain the distance matrix $([\mathbf{D}\mathbf{N}; c \mathbf{1}_{m}^\top] )^2\in \mathbb{R}^{4 \times m}$. Each element in $(\mathbf{D}\mathbf{N})^2\in \mathbb{R}^{3 \times m}$ represents the point-to-line distance between different VPs and lines, as shown in~\cref{fig:DN_example}. Finally, we define the permutation matrix as $\mathbf{Q} \in \mathbb{R}^{4 \times m}$, where each column indicates the association label of each line, as shown in ~\cref{fig:Q_example}. In summary, the primal problem is defined as:
\begin{equation}
    \begin{aligned}
        \min_{\mathbf{D}, \mathbf{Q}} \quad &<([\mathbf{D}\mathbf{N};c \mathbf{1}_{m}^\top] )^2, \mathbf{Q}>  \\
        \text{s.t.} \quad &\mathbf{Q}^{\top}{\mathbf{1}}_4 = {\mathbf{1}}_m,\  (\mathbf{Q})^2 = \mathbf{Q}, \\
        &[\mathbf{D}]_{1,*}[\mathbf{D}]_{1,*}^\top=1, \ [\mathbf{D}]_{2,*}[\mathbf{D}]_{2,*}^\top=1,\\& [\mathbf{D}]_{3,*}[\mathbf{D}]_{3,*}^\top=1, \ 
        [\mathbf{D}]_{1,*}[\mathbf{D}]_{2,*}^\top=0, \\&[\mathbf{D}]_{2,*}[\mathbf{D}]_{3,*}^\top=0, \ [\mathbf{D}]_{1,*}[\mathbf{D}]_{3,*}^\top=0, \\
    \end{aligned}
    \tag{Primal Problem}
    \label{eq:Primal Problem}
\end{equation}
where $c$ denotes the maximum inlier threshold. 

{\it Remarks.} \textbf{1)} The error formulation is designed to be both truncated and multi-selection. ``Truncated'' means that outlier lines are assigned a maximum inlier threshold (\ie, $c^2$). This occurs when the distance between a line and all vanishing points exceeds the maximum inlier threshold (\eg, lines 3 and 4 in~\cref{fig:Q_DN_example}). In this case, the corresponding matrix $\mathbf{Q}$ will label the line as an outlier, set its error to zero, and add a penalty term of $c^2$. ``Multi-selection'' means that we can not only label inliers and outliers, but also identify the vanishing point to which each line belongs. \textbf{2)} The first row of constraints ensures that the elements of the permutation matrix $\mathbf{Q}$ are binary (either ``0'' or ``1''), with exactly single ``1'' in each column. The remaining constraints enforce the orthogonality of the VP matrix $\mathbf{D}$. \textbf{3)} It is clear that when the distance between all outlier lines and all ground-truth vanishing points exceeds the maximum inlier threshold (\ie, $c^2$), the global optimum of \ref{eq:Primal Problem} corresponds to the largest set of inliers and the optimal vanishing point estimation. In this case, the error of the outliers is set to a constant value, ensuring that no bias is introduced.

\subsection{Full QCQP Formulation and Convex Relaxation}
\label{subsec:Full QCQP Formulation and Convex Relaxation}

\subsubsection{Full QCQP Formulation}
Given the primal formulation~\ref{eq:Primal Problem}, we will then derive the equivalent QCQP formulation. To begin with, we firstly introduce a vector $\overline{\mathbf{D}}=[\mathbf{d}_1; \mathbf{d}_2; \mathbf{d}_3; 1] \in \mathbb{R}^{10}$ and a new variable $\boldsymbol{\omega} \in \mathbb{R}^{10(1+4m)}$:
\begin{equation}
    \begin{aligned}
        \boldsymbol{\omega} & =  [\overline{\mathbf{D}};  \text{vec}(\mathbf{Q}) \otimes \overline{\mathbf{D}} ] \\
        & = [\boldsymbol{\omega}_0; \boldsymbol{\omega}_{1,1};\dots;\boldsymbol{\omega}_{4,1};\boldsymbol{\omega}_{1,2};\dots\dots;\boldsymbol{\omega}_{4,m}],
    \end{aligned}
\end{equation}
where 
\begin{equation}
    \left\{\begin{matrix}
    \quad\boldsymbol{\omega}_0 = \overline{\mathbf{D}}\in\mathbb{R}^{10},\quad \boldsymbol{\omega}_{i,j} = [\mathbf{Q}]_{i,j} \overline{\mathbf{D}}\in\mathbb{R}^{10},  \\
    \{\boldsymbol{\omega}_0\}_{i} = \mathbf{d}_i\in\mathbb{R}^{3} \ ( i = 1, 2 ,3 ).
    \end{matrix}\right.
\end{equation}

Additionally, we introduce an auxiliary matrix $\mathbf{C} \in \mathbb{R}^{10(1+4m) \times 10(1+4m)}$, which is defined as a block diagonal matrix: 
\begin{equation} 
\mathbf{C} = \text{diag}([\mathbf{C}_0, \mathbf{C}_{1,1}, \ldots ,\mathbf{C}_{4,1},\mathbf{C}_{1,2}, \ldots\ldots, \mathbf{C}_{4,m}]),
\end{equation}
where
\begin{equation} 
\left\{\begin{matrix}
\mathbf{C}_0 = \mathbf{0}_{10 \times 10}, \\
\{\mathbf{C}_{i,j}\}_{i,i} = \mathbf{n}_j \mathbf{n}_j^{\top}, \quad i = 1,\dots,3, \quad j = 1, \ldots, m, \\
[\mathbf{C}_{4,j}]_{10,10} = c^2, \quad j = 1, \ldots, m.
\end{matrix}\right. 
\end{equation}

Finally, we can equivalently transform \ref{eq:Primal Problem} into the following QCQP problem:
\begin{equation}
    \begin{aligned}
       \min_{\boldsymbol{\omega}} \quad & \boldsymbol{\omega}^{\top} \mathbf{C} \boldsymbol{\omega}\\
        \text{s.t.} \quad &\boldsymbol{\omega}_0 \boldsymbol{\omega}_0^{\top} = \sum_{i=1}^{4} \boldsymbol{\omega}_0 \boldsymbol{\omega}_{i,j}^{\top}, \quad j = 1, \ldots, m, \\
        & \boldsymbol{\omega}_0 \boldsymbol{\omega}_{i,j}^{\top} = \boldsymbol{\omega}_{i,j} \boldsymbol{\omega}_{i,j}^{\top}, \quad \left\{\begin{matrix} i = 1, \ldots, 4 \\ j = 1, \ldots, m \end{matrix}\right., \\ & \{\boldsymbol{\omega}_0\}_{1}^{\top} \{\boldsymbol{\omega}_0\}_{1} = 1, \quad \{\boldsymbol{\omega}_0\}_{2}^{\top} \{\boldsymbol{\omega}_0\}_{2} = 1, \\
        & \{\boldsymbol{\omega}_0\}_{3}^{\top} \{\boldsymbol{\omega}_0\}_{3} = 1, \quad \{\boldsymbol{\omega}_0\}_{1}^{\top} \{\boldsymbol{\omega}_0\}_{2} = 0, \\
        & \{\boldsymbol{\omega}_0\}_{1}^{\top} \{\boldsymbol{\omega}_0\}_{3} = 0, \quad \{\boldsymbol{\omega}_0\}_{2}^{\top} \{\boldsymbol{\omega}_0\}_{3} = 0. \\
    \end{aligned}
    \tag{Full QCQP Problem}
    \label{eq:Full QCQP Problem}
\end{equation}

This QCQP formulation will serve as a bridge to derive the subsequent SDP formulation.

{\it Remark.} The~\ref{eq:Full QCQP Problem} is equivalent to the~\ref{eq:Primal Problem}. In general, the first and second constraints are equivalent to $\mathbf{Q}^{\top}{\mathbf{1}}_4 = {\mathbf{1}}_m$ and $(\mathbf{Q})^2 = \mathbf{Q}$, respectively. The remaining constraints enforce the orthogonality of the VP matrix $\mathbf{D}$.

\subsubsection{Semidefinete Programming Formulation}
Given the definition of~\ref{eq:Full QCQP Problem}, we can relax it to the \ref{eq:Full SDP Problem} as:
\begin{equation}
    \begin{aligned}
        \min_{\mathbf{W}} \quad &  \text{trace}(\mathbf{C} \mathbf{W}) \\ 
        \text{s.t.} \quad &  \mathbf{W}_{0,0} = \sum_{i=1}^{4} \mathbf{W}_{0,4(j-1)+i}, \quad j = 1, \ldots, m,  \\
        \quad & \mathbf{W}_{0,k} = \mathbf{W}_{k,k}, \quad k = 1, \ldots, 4m, \\ 
        \quad & \text{trace}(\{\mathbf{W}_{0,0}\}_{i,j}) = \left\{\begin{matrix} 1,\quad i=j \\ 0,\quad i \neq j \end{matrix}\right. \quad \forall i, j \in \{1, 2, 3\}\\
        \quad &\mathbf{W} \succeq \mathbf{0},
    \end{aligned} 
    \tag{Full SDP Problem}
    \label{eq:Full SDP Problem}
\end{equation}
where 
\begin{equation} 
\begin{aligned}
    \mathbf{W} &\in \mathcal{S}^{10(1+4m)\times10(1+4m)}_{+}  \\
    &= \begin{bmatrix}
  &\mathbf{W}_{0,0} &\mathbf{W}_{0,1} &\dots &\mathbf{W}_{0,4m}& \\
  &\mathbf{W}_{1,0} &\mathbf{W}_{1,1} &\dots &\mathbf{W}_{1,4m} \\
  &\vdots &\vdots &\ddots &\vdots\\
  &\mathbf{W}_{4m,0} &\mathbf{W}_{4m,1} & \dots &\mathbf{W}_{4m,4m}
\end{bmatrix}.
\end{aligned}
\end{equation}

{\it Remark.} The convex relaxation strategy transforms \ref{eq:Full QCQP Problem} into a convex problem, specifically \ref{eq:Full SDP Problem}, by discarding the only non-convex rank constraint $\text{rank}(\mathbf{W}) = 1$. The SDP problem can be globally solved using any off-the-shelf solver in polynomial time. Furthermore, we provide a proof of tight relaxation (\ie, zero duality gap) in \cref{supp:subsec:Full SDP Problem}, which demonstrates that solving \ref{eq:Full SDP Problem} is equivalent to solving \ref{eq:Primal Problem}.

\subsection{GlobustVP: Iterative SDP Solver}
\label{subsec:GlobustVP}

To solve \ref{eq:Full SDP Problem} efficiently, we propose an iterative solver (called {\it \textbf{GlobustVP}}) to estimate the vanishing points one by one. Specifically, in each iteration, we solve~\ref{eq:Single Block SDP Problem} (\cref{subsubsec:Single Block SDP Problem}) to globally search for a single vanishing point, treating the other vanishing points as outliers. In contrast to the previous algorithm, where all three vanishing points were estimated simultaneously, this approach estimates them sequentially. After each iteration, the inlier lines are marked and then excluded from subsequent iterations. After three iterations, \ref{eq:Manhattan Post-Refinement} (\cref{subsubsec:Manhattan Post-Refinement}) is applied to ensure that the vanishing points satisfy the Manhattan world assumption. The algorithm pipeline is summarized in Alg.~\ref{alg:Outlier robust SDP}.

\subsubsection{Single Block SDP Problem}
\label{subsubsec:Single Block SDP Problem}

We first define the tensor $\mathbf{W} \in \mathbb{R}^{3(m+1) \times 3(m+1) \times 2} $ with the following block structure:
\begin{equation} 
\mathbf{W}_{*,*,i} = \begin{bmatrix}
  &\mathbf{W}_{0,0,i} &\mathbf{W}_{0,1,i} &\dots &\mathbf{W}_{0,m,i}& \\
  &\mathbf{W}_{1,0,i} &\mathbf{W}_{1,1,i} &\dots &\mathbf{W}_{1,m,i} \\
  &\vdots &\vdots &\ddots &\vdots\\
  &\mathbf{W}_{m,0,i} &\mathbf{W}_{m,1,i} & \dots &\mathbf{W}_{m,m,i}
\end{bmatrix}.
\end{equation}

Additionally, the auxiliary tensor $\mathbf{C} \in \mathbb{R}^{3(m+1) \times 3(m+1) \times 2}$ is defined as:
\begin{equation}
    \left\{\begin{matrix}
    \mathbf{C}_{*,*,1} = \text{diag}([\mathbf{0}_{3 \times 3}, \mathbf{n}_1 \mathbf{n}_1^{\top}, \ldots, \mathbf{n}_m \mathbf{n}_m^{\top}]), \\
    \mathbf{C}_{*,*,2} = \text{diag}([\mathbf{0}_{3 \times 3}, c^2 \mathbf{I}_3, \ldots, c^2 \mathbf{I}_3]).
    \end{matrix}\right.
    \label{eq:single block sdp C}
\end{equation}

Finally, we convert~\ref{eq:Full SDP Problem} to its iterative variant, \ref{eq:Single Block SDP Problem}, as:
\begin{equation}
    \begin{aligned}
        \min_{\mathbf{W}} \quad &  \text{trace}(\mathbf{C} \mathbf{W}) \\ 
        \text{s.t.} \quad &  \mathbf{W}_{0,0,1} = \sum_{i=1}^{2} \mathbf{W}_{0,j,i}, \quad j = 1, \ldots, m,  \\ \quad & \mathbf{W}_{0,j,i} = \mathbf{W}_{j,j,i}, \quad \forall i \in \{1, 2\}, \quad j = 1, \ldots, m, \\ 
        \quad & \text{trace}(\mathbf{W}_{0,0,1}) = 1, \quad \mathbf{W}_{0,0,1}=\mathbf{W}_{0,0,2},\\
        \quad &\mathbf{W}_{*,*,i}\succeq \mathbf{0}, \quad \forall \ i \in \{1, 2\}.
    \end{aligned} 
     \tag{Single Block SDP Problem}
     \label{eq:Single Block SDP Problem}
\end{equation}

{\it Remark.} \textbf{1)} It is evident that, as long as the distinguishability assumption holds for all lines (\ie, the distance from an inlier line to its corresponding vanishing point is smaller than the maximum inlier threshold, while the distance to other vanishing points is greater than the maximum inlier threshold), \ref{eq:Single Block SDP Problem} will consistently identify the vanishing point with the largest inlier set. \textbf{2)}  In~\cref{supp:subsec:Single Block SDP Problem}, we provide a proof of tight relaxation (\ie, zero duality gap) under noise-free and outlier-free conditions.

\begin{algorithm}[t]
\caption{GlobustVP Solver}
\label{alg:Outlier robust SDP}
\textbf{Input:} A set of detected lines, $\mathbf{n}=\{\mathbf{n}_1, \mathbf{n}_2, \dots, \mathbf{n}_m\}$ \\
\textbf{Output:} Three globally optimal vanishing points and the set of line labels, indicating line associations as VP1, VP2, VP3, Outlier.
\begin{algorithmic}[1]
\FOR{$i = 1:3$}
\STATE Construct the auxiliary tensor $\mathbf{C}_i$ for the line set $\mathbf{n}$ following \cref{eq:single block sdp C},
\STATE Solve the~\ref{eq:Single Block SDP Problem} to obtain $\mathbf{W}_i$,
\STATE Round $\mathbf{W}_i$ and recover one vanishing point, $\mathbf{d}_i$,
\STATE Collect the inlier line IDs in set $\mathbf{l}_i$ and remove these lines from the set $\mathbf{n}$,
\ENDFOR
\STATE Collect the remaining line IDs in set $\mathbf{l}_{\text{outlier}}$,
\STATE Perform the~\ref{eq:Manhattan Post-Refinement} initialized with $\{\mathbf{d}_1, \mathbf{d}_2,\mathbf{d}_3\}$.
\end{algorithmic}
\end{algorithm}

\subsubsection{Manhattan Post-Refinement}
\label{subsubsec:Manhattan Post-Refinement}

After three iterations, given three coarse estimated vanishing points $\{ \mathbf{d}_1, \mathbf{d}_2, \mathbf{d}_3 \}$ and the corresponding set of line labels $\{\mathbf{l}_1, \mathbf{l}_2, \mathbf{l}_3, \mathbf{l}_{\text{outlier}} \}$, we perform~\ref{eq:Manhattan Post-Refinement} to obtain the refined vanishing points $\{ \mathbf{d}_1^*,\mathbf{d}_2^*,\mathbf{d}_3^* \}$:
\begin{equation}
    \begin{aligned}
\min_{\mathbf{d}_1,\mathbf{d}_2,\mathbf{d}_3}\quad& \sum_{i=1}^{3} \sum_{j\in \mathbf{l}_i} (\mathbf{d}_i^{\top}\mathbf{n}_j)^2\\
        \text{s.t.} \quad &\mathbf{d}_1^\top\mathbf{d}_1=1, \mathbf{d}_2^\top\mathbf{d}_2=1, \mathbf{d}_3^\top\mathbf{d}_3=1,\\
        &\mathbf{d}_1^\top\mathbf{d}_2=0, \mathbf{d}_2^\top\mathbf{d}_3=0, \mathbf{d}_1^\top\mathbf{d}_3=0.
    \end{aligned}
    \tag{Manhattan Post-Refinement}
    \label{eq:Manhattan Post-Refinement}
\end{equation}

{\it Remark.} \textbf{1)} Although~\ref{eq:Single Block SDP Problem} does not rely on the Manhattan world assumption, it still yields the optimal inlier line set. When we incorporate the Manhattan world prior into~\ref{eq:Manhattan Post-Refinement}, the final refined vanishing points retain the optimality as in~\ref{eq:Full SDP Problem}. \textbf{2)} It is well known that the optimization time for SDP methods grows cubically with the problem size, making them inefficient for large-scale problems. To overcome this issue, we adopt the strategy from~\cite{le2019sdrsac}, which randomly samples a fixed number of lines for each~\ref{eq:Single Block SDP Problem}. While this aspect is not the main contribution of our work, it provides a highly advantageous trade-off for our algorithm in large-scale settings, as shown in~\cref{fig_line_max_process}.
\section{Experiments}
\label{sec:Experiments}
\subsection{Comparisons against Traditional Methods}
\label{subsec:Comparisons against Traditional Methods}
\subsubsection{Experimental Settings}
\PAR{Baseline methods.} 
We compare our method, {\it \textbf{GlobustVP}}, with the following state-of-the-art traditional approaches:
\begin{itemize}
	\item RANSAC-based method~\cite{Zhang2016VanishingPE} (denoted as \textbf{RANSAC}); 
	\item J-Linkage-based method~\cite{Toldo2008RobustMS} (denoted as \textbf{J-Linkage});
	\item T-Linkage-based method~\cite{Magri2014TLinkageAC} (denoted as \textbf{T-Linkage});
	\item BnB-based method~\cite{Bazin2012GloballyOL} (denoted as \textbf{BnB}); 
	\item RANSAC-BnB hybrid method~\cite{Li2020QuasiGloballyOA} (denoted as \textbf{Quasi-VP}).
\end{itemize}

\PAR{Implementation details.} 
Our approach is implemented in MATLAB, using MOSEK~\cite{MOSEK} as the SDP solver. All experiments are conducted on a laptop with an Intel Core i5-8250U CPU (1.80 GHz) and 8GB of RAM.

\PAR{Evaluation metrics.}
\label{sec:Experiments:Evaluation metrics}

We follow~\cite{Li2020QuasiGloballyOA,Lu20172LineES} to evaluate the accuracy of the image line association in terms of precision $\uparrow$, recall $\uparrow$, and the $F_1$-score $\uparrow$. Specifically, precision is computed as $N_c / (N_c + N_w)$ where $N_c$ and $N_w$ denote the counts of correctly and incorrectly identified inliers, respectively. Recall is defined as $N_c / (N_c + N_m)$ where $N_m$ represents the number of missing inliers. The $F_1$-score, which considers both precision and recall, is given by (2 $\times$ precision $\times$ recall)/(precision+recall). For the evaluation of the real-world dataset, we use the root mean square of the consistency error $\downarrow$ ~\cite{Tardif2009NoniterativeAF,Zhang2016VanishingPE,Li2020QuasiGloballyOA}. This metric is particularly suitable for 3D contexts, as image uncertainties typically arise in the image domain~\cite{Zhang2016VanishingPE}. Specifically, the consistency error quantifies the distance from an endpoint of the image line $l$ to the virtual line $v$, where $v$ is defined by the estimated vanishing point and the midpoint of the line $l$ associated with this vanishing point.

\PAR{Evaluation setup.}
For the synthetic dataset, we generate three distinct sets of parallel 3D lines, ensuring that each pair of lines from different sets is mutually orthogonal. Using a virtual camera, we project these lines onto the virtual image plane, resulting in three sets of image lines. This virtual image has a resolution of $640 \times 480$ pixels and a focal length of 800 pixels, with the principal point located at its center. We then introduce perturbations to the endpoints of these image lines using zero-mean Gaussian noise with a standard deviation of $\sigma$ pixels. Additionally, to generate outlier image lines, we replace a portion of the endpoints with random pixels within the virtual image. The number of lines associated with each vanishing point is randomly distributed, with each vanishing point having a respective number of lines in our experiments. It is important to note that this distribution is not uniform across vanishing points. All statistical analyses are performed across 100 independent Monte Carlo trials, unless otherwise specified.

\begin{figure}[t]
	\centering
	\includegraphics[width=0.99\linewidth]{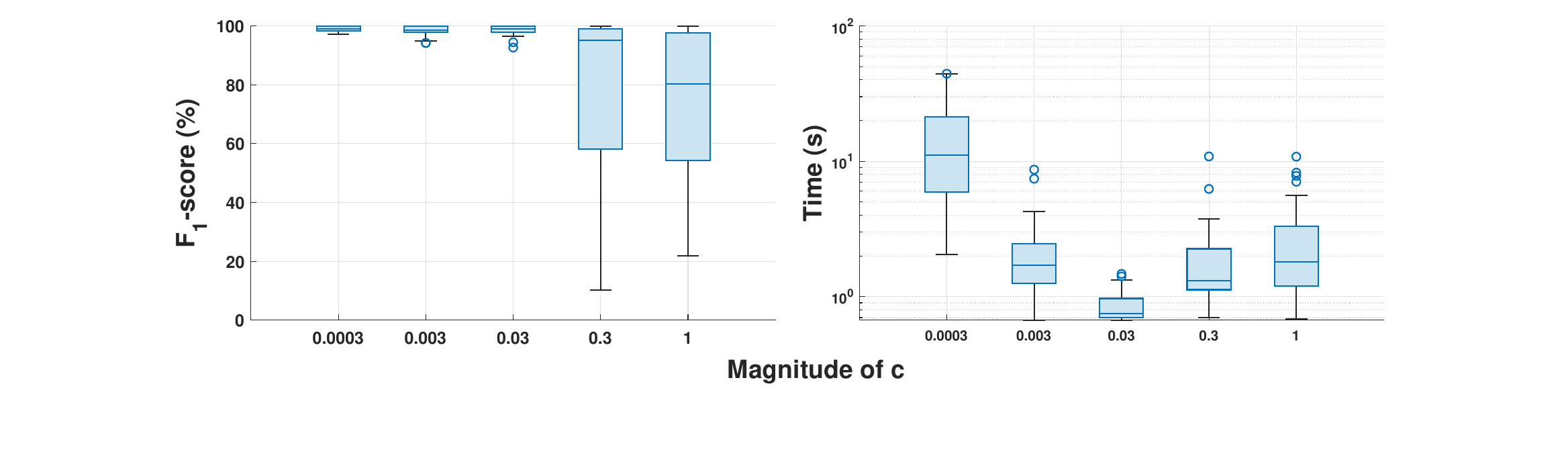}
	\caption{Effect of different values of the maximum inlier threshold $c$. Best viewed in high resolution.}
	\label{fig_ablation_c}
	\vspace{-0.3em}
\end{figure}

\subsubsection{Synthetic Dataset}
\label{subsec:Synthetic Dataset}

\PAR{Effect of $c$.}
To better understand {\it \textbf{GlobustVP}}, we explore different values of $c$ (\ie, maximum inlier threshold introduced in~\cref{subsec:Truncated Multi-selection Error Formulation}), ranging from 0.0003 to 1. We fix the number of lines at 120, the noise level $\sigma$ at 1 pixel, and the outlier ratio at $30\%$. The results in~\cref{fig_ablation_c} show that the $F_1$-score decreases as $c$ increases because larger values of $c$ cause more outliers to be misclassified as inliers, thus reducing the accuracy. Furthermore, the runtime decreases as $c$ increases, reaching a minimum at $c=0.03$, after which it increases again. Therefore, we choose $c=0.03$ for all subsequent experiments, as it provides a good balance between accuracy and computational efficiency.

\begin{figure*}
	\centering
	\includegraphics[width=0.99\linewidth]{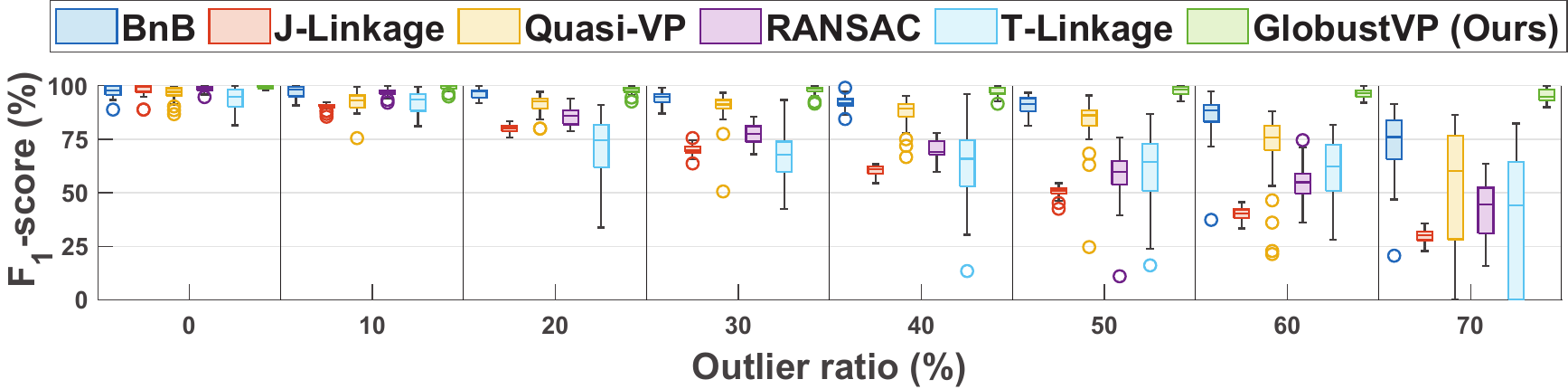}	\caption{Accuracy comparison on the synthetic dataset with respect to the outlier ratios: boxplot of $F_1$-score. Results for precision and recall are provided in \cref{supp:sec:Additional Synthetic Experiments}. Best viewed in color and high resolution.}
	\label{fig_synthetic_accuracy}
\end{figure*}

\PAR{Accuracy comparisons across outlier ratios.}
We use 60 lines, set the noise level at $\sigma = 3$, and increase the outlier ratio from $0\%$ (all inliers) to $70\%$. \cref{fig_synthetic_accuracy} presents $F_1$-score across different outlier ratios. \textbf{RANSAC}, \textbf{J-Linkage}, and \textbf{T-Linkage} maintain robustness only at low outlier ratios, but degrade significantly when outliers exceed $20\%$, while \textbf{Quasi-VP} starts to break down at $40\%$. \textbf{BnB} generally handles high outlier ratios well, while occasional convergence issues reduce its accuracy. In contrast, {\it \textbf{GlobustVP}} consistently demonstrates superior robustness and accuracy across varying outlier ratios. The complete results, including precision and recall, are provided in \cref{supp:sec:Additional Synthetic Experiments}.

\begin{figure}[t]
	\centering
	\includegraphics[width=0.99\linewidth]{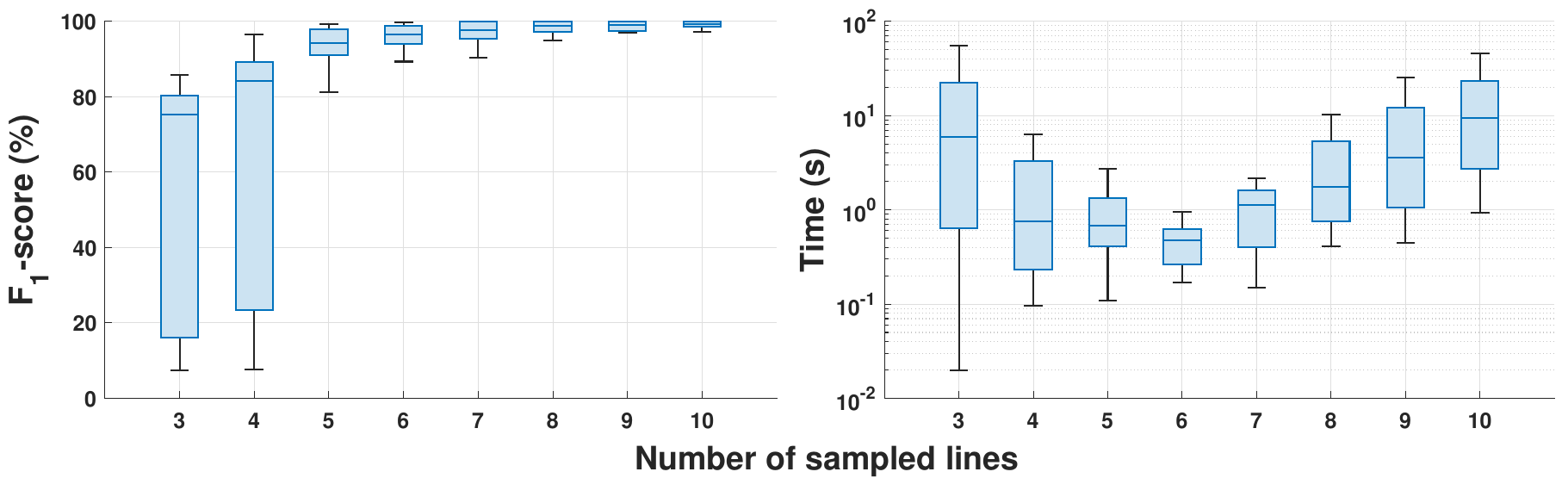}
	\caption{Accuracy and efficiency comparisons with respect to the number of sampled lines on the synthetic dataset. Best viewed in high resolution.}
	\label{fig_line_max_process}
	\vspace{0.75em}
\end{figure}

\PAR{Sampling acceleration in large-scale scenario.}
To evaluate sampling acceleration in large-scale scenarios, we generate $120$ lines, setting the noise level at $\sigma = 3$ and the outlier ratio at $20\%$. We vary the number of sampled lines from 3 to 10, with an interval of 1. \cref{fig_line_max_process} shows both the $F_1$-score and runtime. The total runtime decreases as the number of sampled lines increases, reaching a minimum at 6, and then increases again. In contrast, the accuracy (\ie, $F_1$-score) improves as the number of sampled lines grows, stabilizing at 6 sampled lines and remaining consistent for larger values. Therefore, in the following efficiency experiments, we fix the number of sampled lines at 6, which provides the optimal trade-off in our method.

\begin{figure}[t]
	\centering
	\includegraphics[width=0.99\linewidth]{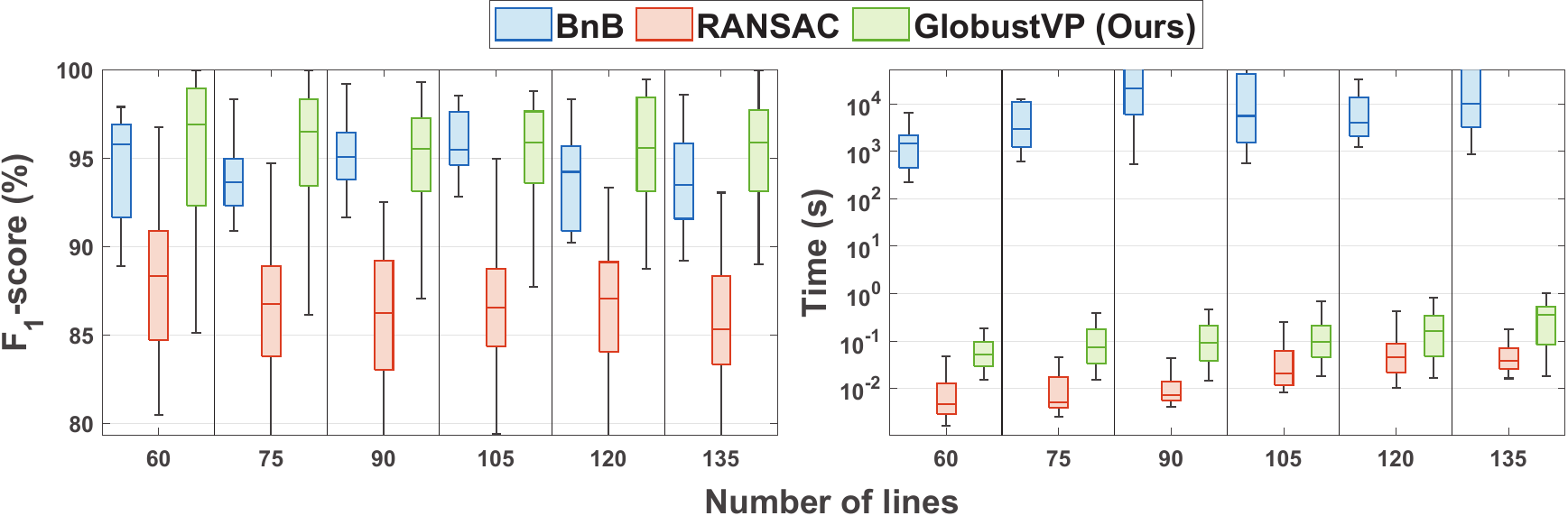}
	\caption{Accuracy and efficiency comparisons of \textbf{RANSAC}~\cite{Zhang2016VanishingPE} and \textbf{BnB}~\cite{Bazin2012GloballyOL} on the synthetic dataset with respect to the number of image lines. Complete comparisons of all baseline methods are provided in \cref{supp:sec:Additional Synthetic Experiments}. Best viewed in color and high resolution.}
	\label{fig_line_number_ransac_bnb_ours}
\end{figure}

\PAR{Accuracy and efficiency comparisons across line counts.}
Using the sampling acceleration strategy described above, where the optimal number of sampled lines is determined to be 6, we conduct a thorough assessment of accuracy and efficiency with respect to the number of image lines. Specifically, we set the noise level at $\sigma = 3$ and the outlier ratio at $20\%$, while varying the number of image lines from 60 to 135. For clarity, we present comparisons here specifically with \textbf{RANSAC} and \textbf{BnB}, while complete results of all the baseline methods are provided in \cref{supp:sec:Additional Synthetic Experiments}. As shown in \cref{fig_line_number_ransac_bnb_ours}, we report both the $F_1$-score and the runtime as the number of image lines increases. While \textbf{RANSAC} achieves the highest efficiency, it suffers from a lower accuracy (\ie, $F_1$-score). In contrast, {\it \textbf{GlobustVP}} achieves significantly faster runtime compared to \textbf{BnB} while maintaining robust accuracy. This balance between efficiency and accuracy highlights the practical applicability of our method in real-world scenarios.

\subsubsection{Real-World Dataset}
\begin{figure*}[ht]
	\centering
	\bigskip
	\begin{subfigure}[b]{0.24\textwidth}
		\centering
		\caption*{\centering Lines}
		\includegraphics[width=\textwidth]{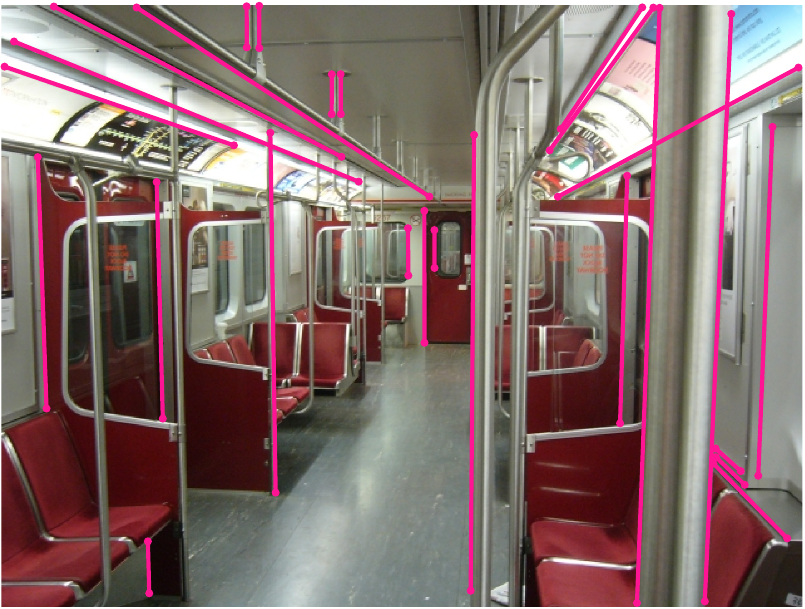}
		\caption*{\centering 3 VPs \par 32 lines}
		\label{fig_real_vis_manu_3vp_lines}
	\end{subfigure}
	\begin{subfigure}[b]{0.24\textwidth}
		\centering
		\caption*{\centering Ground Truth}
		\includegraphics[width=\textwidth]{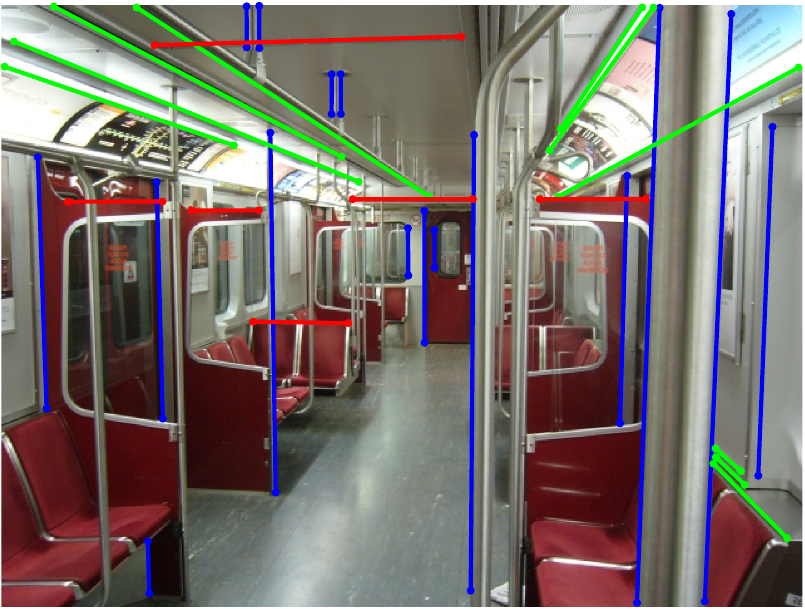}
		\caption*{\centering 3 VPs \par 32 lines}
		\label{fig_real_vis_manu_3vp_gt}
	\end{subfigure}
	\begin{subfigure}[b]{0.24\textwidth}
		\centering
		\caption*{\centering \textbf{RANSAC}~\cite{Zhang2016VanishingPE}}
		\includegraphics[width=\textwidth]{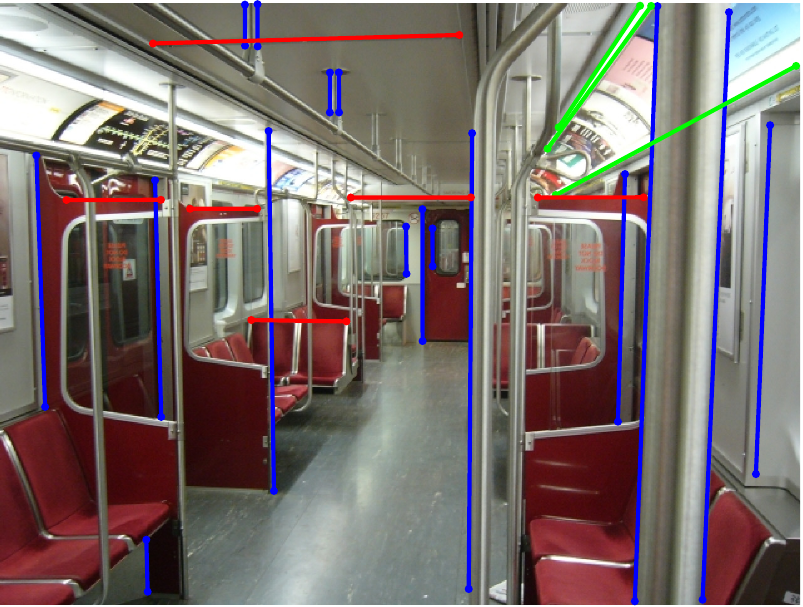}
		\centering
		\caption*{\centering \textcolor{pink}{$100\%$}, \textcolor{purple}{$75.00\%$} \par \textcolor{teal}{$85.71\%$}, \textcolor{violet}{1.01 pix.}}
		\label{fig_real_vis_manu_3vp_ransac}
	\end{subfigure}
	\begin{subfigure}[b]{0.24\textwidth}
		\centering
		\caption*{\centering \textbf{J-Linkage}~\cite{Toldo2008RobustMS}}
		\includegraphics[width=\textwidth]{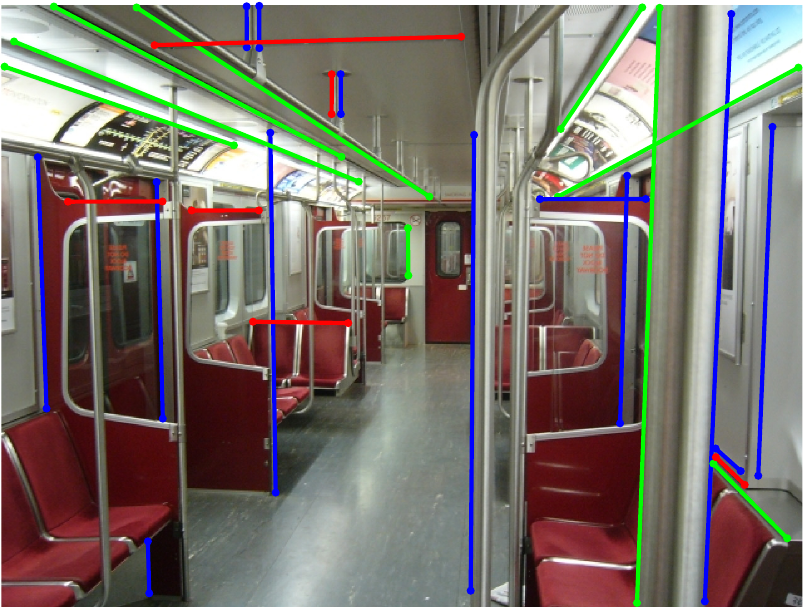}
		\caption*{\centering \textcolor{pink}{$78.57\%$}, \textcolor{purple}{$84.61\%$} \par \textcolor{teal}{$81.48\%$}, \textcolor{violet}{1.24 pix.}}
		\label{fig_real_vis_manu_3vp_jlinkage}
	\end{subfigure}
	
	\begin{subfigure}[b]{0.24\textwidth}
		\centering
		\caption*{\centering \textbf{T-Linkage}~\cite{Magri2014TLinkageAC}}
		\includegraphics[width=\textwidth]{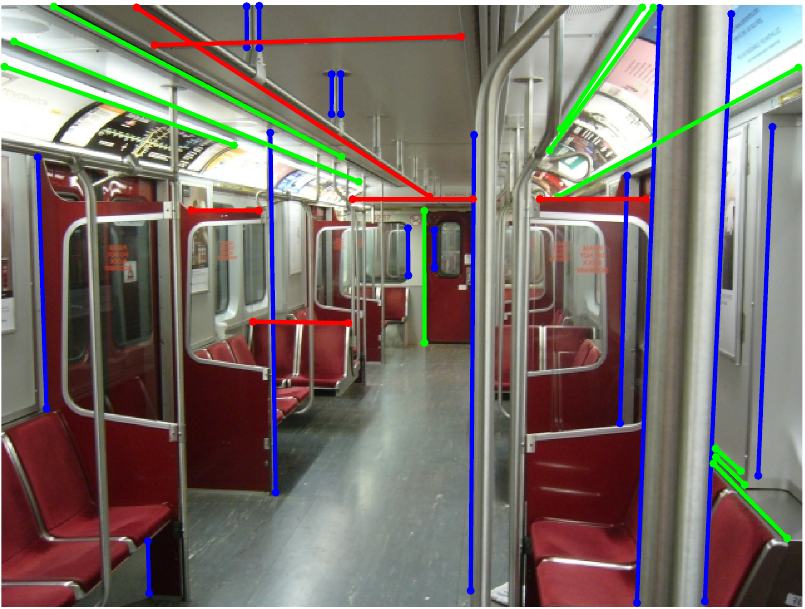}
		\caption*{\centering \textcolor{pink}{$93.33\%$}, \textcolor{purple}{$93.33\%$} \par \textcolor{teal}{93.33\%}, \textcolor{violet}{0.63 pix.}}
		\label{fig_real_vis_manu_3vp_tlinkage}
	\end{subfigure}
	\begin{subfigure}[b]{0.24\textwidth}
		\centering
		\caption*{\centering \textbf{BnB}~\cite{Bazin2012GloballyOL}}
		\includegraphics[width=\textwidth]{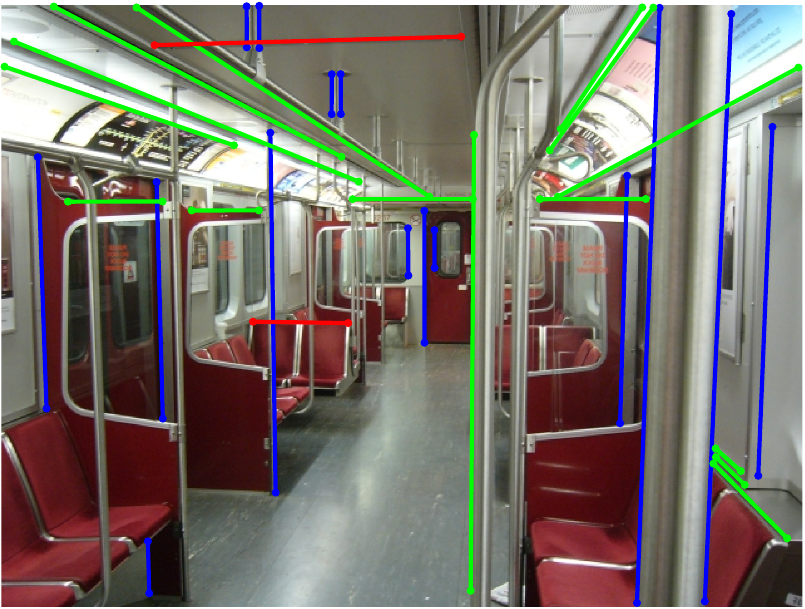}
		\caption*{\centering \textcolor{pink}{$64.29\%$}, \textcolor{purple}{$64.29\%$} \par \textcolor{teal}{$64.29\%$}, \textcolor{violet}{2.19 pix.}}
		\label{fig_real_vis_manu_3vp_bnb}
	\end{subfigure}
	\begin{subfigure}[b]{0.24\textwidth}
		\centering
		\caption*{\centering \textbf{Quasi-VP}~\cite{Li2020QuasiGloballyOA}}
		\includegraphics[width=\textwidth]{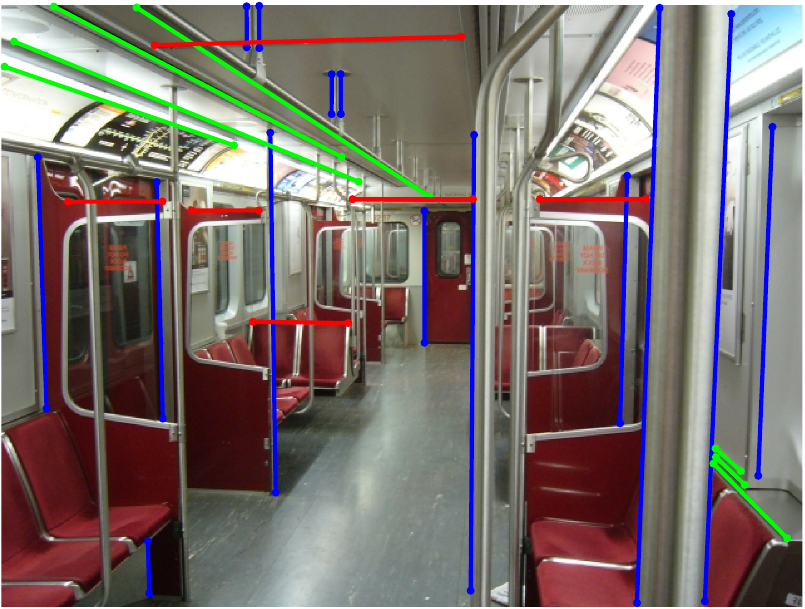}
		\caption*{\centering \textcolor{pink}{$100\%$}, \textcolor{purple}{$64.29\%$} \par \textcolor{teal}{$78.26\%$}, \textcolor{violet}{3.24 pix.}}
		\label{fig_real_vis_manu_3vp_quasi}
	\end{subfigure}
	\begin{subfigure}[b]{0.24\textwidth}
		\centering
		\caption*{\centering \textbf{GlobustVP (Ours)}}
		\includegraphics[width=\textwidth]{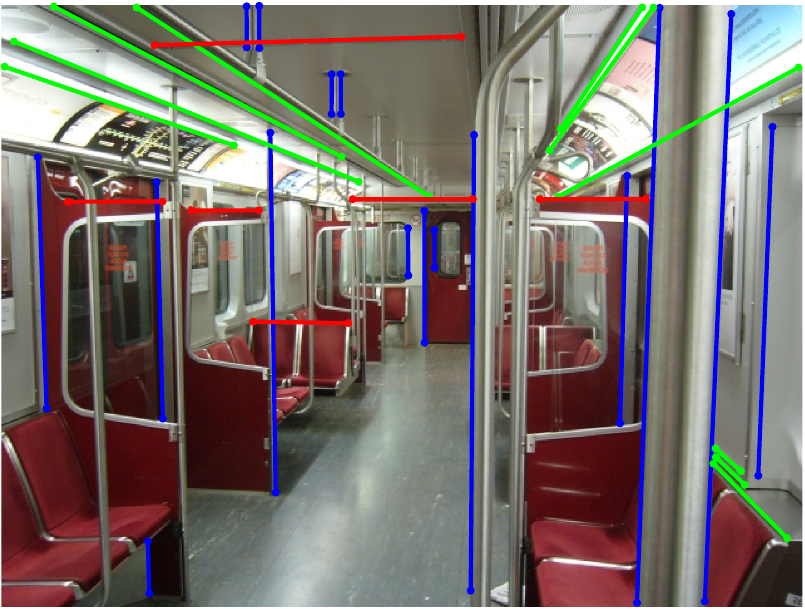}
		\caption*{\centering \textcolor{pink}{\textbf{$100\%$}}, \textcolor{purple}{\textbf{$100\%$}} \par \textcolor{teal}{\textbf{$100\%$}}, \textcolor{violet}{\textbf{0.41 pix.}}}
		\label{fig_real_vis_manu_3vp_ours}
	\end{subfigure}
	\caption{Representative comparisons on YUD~\cite{Denis2008EfficientEM} using the manually extracted image lines. Different line-VP associations are shown in respective colors. The numbers below each image represent the respective \textcolor{pink}{precision $\uparrow$}, \textcolor{purple}{recall $\uparrow$}, \textcolor{teal}{$F_1$-score $\uparrow$}, and \textcolor{violet}{consistency error $\downarrow$} of line-VP association. Best viewed in color and high resolution.}
	\label{fig_real_vis_manu}
\end{figure*}

To evaluate different methods on real-world images, we conduct experiments using the York Urban Database (YUD)~\cite{Denis2008EfficientEM}. This dataset is widely used for evaluating the estimation of vanishing points and the association of lines in the Manhattan world. It consists of 102 calibrated images, each with a resolution of $640 \times 480$ pixels, taken from both indoor and outdoor environments. Each image contains a set of manually annotated inlier line segments (all inliers), corresponding to either 2 or 3 vanishing points, along with the ground truth line-VP associations.
\cref{fig_teaser,fig_real_vis_manu} show line-VP association results obtained by {\it \textbf{GlobustVP}} and other state-of-the-art techniques.
A detailed analysis of $F_1$-score and consistency error on all images of YUD~\cite{Denis2008EfficientEM}, along with further quantitative comparisons, is provided in \cref{supp:subsec:York Urban Database}.

\begin{table}[t]
	\centering
	\resizebox{0.99\linewidth}{!}{
		\begin{tabular}{l ccc ccc}
			\toprule
			Datasets & \multicolumn{3}{c}{YUD~\cite{Denis2008EfficientEM}} & \multicolumn{3}{c}{SU3~\cite{zhou2019learning}}\\ 
			\cmidrule(l){2-4} \cmidrule(l){5-7}
			Metrics & AA@$3^{\circ}$ $\uparrow$ & AA@$5^{\circ}$ $\uparrow$ & AA@$10^{\circ}$ $\uparrow$ & AA@$3^{\circ}$ $\uparrow$ & AA@$5^{\circ}$ $\uparrow$ & AA@$10^{\circ}$ $\uparrow$ \\  
			\midrule
			J-Linkage~\cite{Toldo2008RobustMS} & 57.7 & 69.3 & 80.5 & 70.2 & 77.6 & 83.8 \\
			Contrario-VP~\cite{simon2018contrario} & 58.3 & 71.1 & 80.7 & 66.8 & 74.0 & 81.6 \\
			Quasi-VP~\cite{Li2020QuasiGloballyOA} & 57.8 & 72.5 & 84.3 & 72.2 & 78.8 & 81.8 \\
			\midrule
			Zhai~\etal~\cite{zhai2016detecting} & 48.9 & 63.1 & 76.5 & 65.4 & 75.4 & 82.2 \\
			NeurVPS~\cite{Zhou2019NeurVPSNV} & 52.2 & 64.2 & 78.1 & \cellcolor{tabfirst}93.2 & \cellcolor{tabfirst}95.2 & \cellcolor{tabfirst}97.8 \\
			CONSAC~\cite{Kluger2020CONSACRM} & 61.9 & 73.5 & 83.4 & 79.2 & 85.8 & 90.6 \\
			Lin~\etal~\cite{lin2022deep}  & 63.3 & 75.6 & 85.7 & 84.2 & 90.4 & 95.0 \\
			PARSAC~\cite{kluger2024parsac} & \cellcolor{tabsecond}64.7 & \cellcolor{tabsecond}77.2 & \cellcolor{tabsecond}85.6 & \cellcolor{tabsecond}86.2 & \cellcolor{tabsecond}90.6 & \cellcolor{tabsecond}95.4 \\
			\midrule
			GlobustVP (Ours) & \cellcolor{tabfirst}\textbf{67.6} & \cellcolor{tabfirst}\textbf{87.3} & \cellcolor{tabfirst}\textbf{96.1} & \textbf{80.2} & \textbf{86.8} & \textbf{92.4}\\
			\bottomrule
		\end{tabular}
	}
	\caption{Angular accuracy on YUD~\cite{Denis2008EfficientEM} and SU3~\cite{zhou2019learning} datasets. {\it \textbf{GlobustVP}} achieves the best results on YUD~\cite{Denis2008EfficientEM}, and is competitive on SU3~\cite{zhou2019learning}. The \colorbox{tabfirst}{best} and \colorbox{tabsecond}{second-best} performance for each metric are highlighted.}
	\label{tab_learning_eval}
	\vspace{-0.1em}
\end{table}

\subsection{Comparisons against Learning-Based Methods}
\label{sec:Comparisons with learning-based methods}
While our method outperforms traditional approaches on both synthetic and real data, it is worth noting that learning-based methods have recently demonstrated promising results in vanishing point estimation.

\subsubsection{Experimental Settings}
\label{subsubsec:Real Experimental Settings}

\PAR{Datasets.}
In addition to YUD~\cite{Denis2008EfficientEM}, we also evaluate on the SceneCity Urban 3D (SU3) wireframe dataset~\cite{zhou2019learning}. The SU3 dataset consists of 23K synthetic images, divided into $80\%$ for training, $10\%$ for validation, and $10\%$ for testing.

\PAR{Baseline methods.} 
We compare our method, {\it \textbf{GlobustVP}} with the state-of-the-art learning-based methods and traditional, including~\cite{zhai2016detecting}, \textbf{NeurVPS}~\cite{Zhou2019NeurVPSNV}, \textbf{CONSAC}~\cite{Kluger2020CONSACRM},~\cite{lin2022deep}, and \textbf{PARSAC}~\cite{kluger2024parsac} on YUD~\cite{Denis2008EfficientEM} and SU3~\cite{zhou2019learning} datasets.

\PAR{Evaluation protocol.}
Following~\cite{lin2022deep}, we train~\cite{zhai2016detecting} and \textbf{CONSAC} from scratch on SU3~\cite{zhou2019learning}, while using official code and pre-trained models for \textbf{NeurVPS}, \cite{lin2022deep}, and \textbf{PARSAC}, which are typically well-trained on SU3~\cite{zhou2019learning}.
Additionally, we evaluate traditional methods, including \textbf{J-Linkage}, \textbf{Contrario-VP}~\cite{simon2018contrario}, and \textbf{Quasi-VP}.
\textbf{Contrario-VP} employs an a-contrario framework to constrain vanishing points on the horizon line. 
To ensure consistency, we follow~\cite{Zhou2019NeurVPSNV} and adopt the angle accuracy (AA) metric, which measures the angle between the estimated and ground truth dominant directions. We then calculate the percentage of estimates with angular differences below predefined thresholds (3$^{\circ}$, 5$^{\circ}$, and 10$^{\circ}$).

\subsubsection{Results}
As shown in~\cref{tab_learning_eval}, {\it \textbf{GlobustVP}} achieves the highest AA scores on YUD~\cite{Denis2008EfficientEM} across all thresholds, attaining an AA@$10^{\circ}$ of $96.1\%$, which significantly outperforms both learning-based and traditional approaches. On SU3~\cite{zhou2019learning}, {\it \textbf{GlobustVP}} demonstrates competitive performance, particularly compared to traditional baselines, while learning-based methods, such as \textbf{NeurVPS} and \textbf{PARSAC}, achieve better results due to their training on similar data.
This difference in generalization capability highlights the robustness of {\it \textbf{GlobustVP}} in diverse settings, as it does not rely on data-specific training. By effectively balancing accuracy, robustness, and generalization, {\it \textbf{GlobustVP}} proves well-suited for real-world applications where data-driven methods often struggle with limited generalization.
\section{Conclusions}
\label{sec:Conclusion}

In this paper, we introduce convex relaxation techniques to VP estimation for the first time.
Specifically, we propose a ``soft'' association scheme that allows for jointly inferring the line-VP association and locating each VP.
The core for establishing such a scheme is the proposed truncated multi-selection error, an attribute to which the original primal problem can be reformulated to a QCQP form and further relaxed to a convex SDP problem.
Besides, we propose an iterative solver, \textbf{\textit{GlobustVP}}, for efficiently solving the resulting SDP problem, enjoying efficient computation while preserving global optimality under mild conditions. 
Experimental results justify the effectiveness of our algorithm, outperforming existing methods in terms of efficiency, robustness to outliers, and guaranteed global optimality.

\section*{Acknowledgments}
This work is supported by the National Natural Science Foundation of China (62202389, 62403401), in part by a grant from the Westlake University-Muyuan Joint Research Institute, and in part by the Westlake Education Foundation.

\appendix
\section*{Appendix}
\label{sec:Appendix}

\noindent In the Appendix, we provide the following contents:
\begin{itemize}
    \item Theoretical proof of the global optimality of our proposed solvers in \cref{subsec:Full QCQP Formulation and Convex Relaxation,subsec:GlobustVP} (\cref{supp:sec:Theoretical Proof}).
    \item Rounding and nearest rotation recovery implementation (\cref{supp:sec:Rounding and Nearest Rotation Recovery}).
    \item Additional experiment analysis of our solver on the synthetic dataset (\cref{supp:sec:Additional Synthetic Experiments}).
    \item Analysis of different RANSAC-based methods (\cref{supp:sec:Analysis of RANSAC-based methods}).
    \item Additional experiment analysis of our solver on the real-world dataset (\cref{supp:sec:Additional Real-World Experiments}).
\end{itemize}

\section{Theoretical Proof}
\label{supp:sec:Theoretical Proof}
In this section, we provide a theoretical proof of two proposed SDP solvers.

\subsection{Preliminary Lemma}
\noindent To begin with, define the multi block primal QCQP as:
\begin{equation}
    \begin{aligned}
        \min_{\mathbf{x}_i \in \mathbb{R}^n} \quad & \sum_{i=1}^{l}  \mathbf{x}^{\top}_i \mathbf{C}_i \mathbf{x}_i\\
         \text{s.t. } \quad & \sum_{i=1}^{l} \mathbf{x}^{\top}_i \mathbf{A}_{ij} \mathbf{x}_i = b_j, \quad j = 1, \ldots, m.
    \end{aligned}
    \tag{Primal QCQP}
    \label{supp:eq:Primal QCQP}
\end{equation}
\noindent The Lagrange dual of primal QCQP can be derived as:
\begin{equation}
    \begin{aligned}
        \max_{\mathbf{y} \in \mathbb{R}^m} \quad & \mathbf{b}^{\top}\mathbf{y}\\
         \text{s.t. } \quad &  \mathbf{C}_i - \sum_{j = 1}^{m} y_j\mathbf{A}_{ij}  \succeq 0, \quad i = 1, \ldots, l.
    \end{aligned}
    \tag{Dual SDP}
    \label{supp:eq:Dual SDP}
\end{equation}
\noindent The dual of Lagrange dual of primal QCQP can be derived as:
\begin{equation}
    \begin{aligned}
        \min_{\mathbf{X}_i \in \mathcal{S}^n} \quad & \sum_{i=1}^{l} \text{trace}(\mathbf{C}_i\mathbf{X}_i)\\
         \text{s.t. } \quad &  \sum_{i=1}^{l} \text{trace}(\mathbf{A}_{ij}\mathbf{X}_i) = b_j, \quad j = 1, \ldots, m\\
         & \mathbf{X}_i \succeq \mathbf{0}, \quad i = 1, \ldots, l.
    \end{aligned}
    \tag{Dual Dual SDP}
    \label{supp:eq:Dual Dual SDP}
\end{equation}

\noindent Based on Lemma 2.1 in \cite{cifuentes2022local}, we can define the following lemmas: 

~\\
\noindent\textbf{Lemma 1.} Let $\mathcal{H}(\mathbf{y})_i =\mathbf{C}_i - \sum_{j = 1}^{m} y_j\mathbf{A}_{ij}$. $\mathbf{x}$ is proved to be optimal for problem \ref{supp:eq:Primal QCQP}, the strong duality holds between \ref{supp:eq:Primal QCQP} and \ref{supp:eq:Dual SDP} if there exists $\mathbf{x}_i \in \mathbb{R}^{n}, \mathbf{y} \in \mathbb{R}^m$ satisfies:
\begin{equation}
    \left\{\begin{matrix}
    \sum_{i=1}^{l} \mathbf{x}^{\top}_i \mathbf{A}_{ij} \mathbf{x}_i = b_j, \ j = 1, \ldots, m   &(\text{Primal Feasibility})\\
    \mathcal{H}(\mathbf{y})_i \succeq 0, \  i = 1, \ldots, l &(\text{Dual Feasibility})\\
    \mathcal{H}(\mathbf{y})_i\mathbf{x}_i = 0, \  i = 1, \ldots, l &(\text{Stationary Condition})
    \end{matrix}\right.
\end{equation}

~\\
\noindent\textbf{Lemma 2.} In addition to Lemma 1, if $\mathcal{H}(\mathbf{y})_i$ has corank one, then all $\mathbf{x}_i\mathbf{x}_i^{\top}$ are the unique optimum of \ref{supp:eq:Dual Dual SDP} and all $\mathbf{x}_i$ are the unique optimum of \ref{supp:eq:Primal QCQP}.
~\\

\noindent Based on these two lemmas, we can prove the strong duality of our methods as follows. 

\subsection{Full SDP Problem}
\label{supp:subsec:Full SDP Problem}
Let us first revisit \ref{eq:Full SDP Problem} as:
\begin{equation}
    \begin{aligned}
        \min_{\mathbf{W}} \quad &  \text{trace}(\mathbf{C} \mathbf{W}) \\ 
        \text{s.t.} \quad &  \mathbf{W}_{0,0} = \sum_{i=1}^{4} \mathbf{W}_{0,4(j-1)+i}, \quad j = 1, \ldots, m,  \\
        \quad & \mathbf{W}_{0,k} = \mathbf{W}_{k,k}, \quad k = 1, \ldots, 4m, \\ 
        \quad & \text{trace}(\{\mathbf{W}_{0,0}\}_{i,j}) = \left\{\begin{matrix} 1,\quad i=j \\ 0,\quad i\ne j \end{matrix}\right., \quad \forall i,j\in \{1, 2, 3\},\\
        \quad &\mathbf{W} \succeq \mathbf{0},
    \end{aligned} 
    \tag{Full SDP Problem}
\end{equation}
where 
\begin{equation} 
\begin{aligned}
    \mathbf{W} &\in \mathcal{S}^{10(1+4m)\times10(1+4m)}_{+}  \\
    &= \begin{bmatrix}
  &\mathbf{W}_{0,0} &\mathbf{W}_{0,1} &\dots &\mathbf{W}_{0,4m}& \\
  &\mathbf{W}_{1,0} &\mathbf{W}_{1,1} &\dots &\mathbf{W}_{1,4m} \\
  &\vdots &\vdots &\ddots &\vdots\\
  &\mathbf{W}_{4m,0} &\mathbf{W}_{4m,1} & \dots &\mathbf{W}_{4m,4m} 
\end{bmatrix}. 
\end{aligned}
\end{equation}

~\\
\noindent\textbf{Theorem 1.} The duality gap of \ref{eq:Full SDP Problem} is zero under the noise-free and outlier-free condition. The \ref{eq:Full SDP Problem} solver guarantees that the optimal solution with rank 1.

~\\
\textit{Proof: } Let we simplify the \ref{eq:Full SDP Problem} as: 
\begin{equation}
    \begin{aligned}
        \min_{\mathbf{W}} \quad &  \text{trace}(\mathbf{C} \mathbf{W}) \\ 
        \text{s.t.} \quad &  \text{trace}(\mathbf{A}_1 \mathbf{W})=0, \quad j = 1, \ldots, m,  \\
        \quad & \text{trace}(\mathbf{A}_2 \mathbf{W})=0, \quad k = 1, \ldots, 4m, \\ 
        \quad & \text{trace}(\mathbf{A}_3\mathbf{W}) = 0,\\ 
        \quad & \text{trace}(\mathbf{A}_4\mathbf{W}) = 1, \quad \text{trace}(\mathbf{A}_5\mathbf{W}) = 1, \\ 
        \quad &\text{trace}(\mathbf{A}_6\mathbf{W}) = 1,\\
        \quad &\mathbf{W} \succeq \mathbf{0},
    \end{aligned} 
    \tag{Full SDP Problem}
\end{equation}
where the corresponding Lagrange multipliers defined as:
\begin{equation}
\mathbf{y}=\left[\mathbf{y}_1\in\mathbb{R}^m;\mathbf{y}_2\in\mathbb{R}^{4m};y_3;y_4;y_5;y_6\right].
\end{equation}
Given the ground truth vanishing points $[\mathbf{d}_1^*;\mathbf{d}_2^*;\mathbf{d}_3^*]$ and permutation matrix $\mathbf{Q}^*$, we can derive the optimal $\boldsymbol{\omega}^*$ as:
\begin{equation}
    \boldsymbol{\omega}^* =  [\overline{\mathbf{D}}^*;  \text{vec}(\mathbf{Q}^*) \otimes \overline{\mathbf{D}}^* ],
\end{equation}
where $\overline{\mathbf{D}}^*=[\mathbf{d}_1^*;\mathbf{d}_2^*;\mathbf{d}_3^*;1]$ represents the homogeneous vector. Then, let $\mathbf{y}=\mathbf{0}$ become a zero vector. This implies that $\mathcal{H}(\mathbf{y})=\mathbf{C}$ and $\boldsymbol{\omega}^{*\top}\mathcal{H}(\mathbf{y})\boldsymbol{\omega}^*=\boldsymbol{\omega}^{*\top}\mathbf{C}\boldsymbol{\omega}^*$.

Since $\boldsymbol{\omega}^*$ satisfies our primal feasibility and $\mathbf{C}$ is diagonally symmetric, which implies $\mathcal{H}(\mathbf{y}) \succeq 0$. In addition, $\boldsymbol{\omega}^{*\top}\mathbf{C}\boldsymbol{\omega}^*=0 \Longrightarrow \mathbf{C}\boldsymbol{\omega}^*=\mathbf{0}$. Thus the relaxation is tight according to Lemma 1. Since $\boldsymbol{\omega}$ is the only nonzero solution to $\boldsymbol{\omega}^{\top}\mathcal{H}(\mathbf{y})\boldsymbol{\omega}=0$ up to scale, according to our Lemma 2, our \ref{eq:Full SDP Problem} can always return the optimal solution. $\hfill\square$

\subsection{Single Block SDP Problem}
\label{supp:subsec:Single Block SDP Problem}
Let us revisit \ref{eq:Single Block SDP Problem} as:
\begin{equation}
    \begin{aligned}
        \min_{\mathbf{W}} \quad &  \text{trace}(\mathbf{C} \mathbf{W}) \\ 
        \text{s.t.} \quad &  \mathbf{W}_{0,0,1} = \sum_{i=1}^{2} \mathbf{W}_{0,j,i}, \quad j = 1, \ldots, m,  \\ \quad & \mathbf{W}_{0,j,i} = \mathbf{W}_{j,j,i}, \quad \forall i \in \{1, 2\}, \quad j = 1, \ldots, m,  \\ 
        \quad & \text{trace}(\mathbf{W}_{0,0,1}) = 1, \quad \mathbf{W}_{0,0,1}=\mathbf{W}_{0,0,2},\\
        \quad &\mathbf{W}_{*,*,i}\succeq \mathbf{0}, \quad \forall i \in \{1, 2\},
    \end{aligned} 
     \tag{Single Block SDP Problem}
\end{equation}
where the tensor $\mathbf{W}\in \mathbb{R}^{3(m+1) \times 3(m+1) \times 2} $ with its block structure is defined as:
\begin{equation} 
\mathbf{W}_{*,*,i} = \begin{bmatrix}
  &\mathbf{W}_{0,0,i} &\mathbf{W}_{0,1,i} &\dots &\mathbf{W}_{0,m,i}& \\
  &\mathbf{W}_{1,0,i} &\mathbf{W}_{1,1,i} &\dots &\mathbf{W}_{1,m,i} \\
  &\vdots &\vdots &\ddots &\vdots\\
  &\mathbf{W}_{m,0,i} &\mathbf{W}_{m,1,i} & \dots &\mathbf{W}_{m,m,i}
\end{bmatrix}.
\end{equation}
Besides, the auxiliary tensor $\mathbf{C} \in \mathbb{R}^{3(m+1) \times 3(m+1) \times 2}$ is defined as:
\begin{equation}
    \left\{\begin{matrix}
    \mathbf{C}_{*,*,1} = \text{diag}([\mathbf{0}_{3 \times 3}, \mathbf{n}_1 \mathbf{n}_1^{\top}, \ldots, \mathbf{n}_m \mathbf{n}_m^{\top}])\\
    \mathbf{C}_{*,*,2} = \text{diag}([\mathbf{0}_{3 \times 3}, c^2 \mathbf{I}_3, \ldots, c^2 \mathbf{I}_3])
    \end{matrix}\right.
\end{equation}

~\\
\noindent \textbf{Theorem 2.} The duality gap of \ref{eq:Single Block SDP Problem} is zero under the noise-free and outlier-free condition. The \ref{eq:Single Block SDP Problem} solver guarantees that the optimal solution with rank 1.

~\\
\textit{Proof: } Let we simplify the single block QCQP as:
\begin{equation}
    \begin{aligned}
         (\mathbf{x}_1^{*},\mathbf{x}_2^{*}) = arg \min_{\mathbf{x}_1,\mathbf{x}_2} \quad &  \mathbf{x}_1^{\top}\mathbf{C}_1\mathbf{x}_1 + \mathbf{x}_2^{\top}\mathbf{C}_2\mathbf{x}_2 \\ 
        \text{s.t. } \quad & \mathbf{x}^{\top}_1 \mathbf{A}_{1j} \mathbf{x}_1 = b_{1j}, \quad j = 1, \ldots, m1, \\
        & \mathbf{x}^{\top}_2 \mathbf{A}_{2j} \mathbf{x}_2 = b_{2j}, \quad j = 1, \ldots, m2, \\
        & \mathbf{x}^{\top}_1 \mathbf{A}_{3j1} \mathbf{x}_1 + \mathbf{x}^{\top}_2 \mathbf{A}_{3j2} \mathbf{x}_2 = b_{3j}, \\
        & \quad j = 1, \ldots, m3.
    \end{aligned}
\end{equation}

After setting all Lagrange multipliers to zero, we have $\mathcal{H}(\mathbf{y})_1 = \mathbf{C}_1$ and $\mathcal{H}(\mathbf{y})_2 = \mathbf{C}_2$, respectively. Since we assume that all lines are inliers, the ground truth $\mathbf{x}_1$ and $\mathbf{x}_2$ can be defined as:
\begin{equation}
    \begin{aligned}
         \mathbf{x}^{*}_1 &= [\mathbf{d}^{*};\mathbf{d}^{*};\ldots;\mathbf{d}^{*}], \\
         \mathbf{x}^{*}_2 &= [\mathbf{d}^{*};\mathbf{0};\ldots;\mathbf{0}],
    \end{aligned}
\end{equation}
where $\mathbf{d}^{*}$ denotes the ground truth 3D line direction.  The primal feasibility is always satisfied and the dual feasibility can be verified as:
\begin{equation}
    \begin{aligned}
        0 &= \mathbf{x}_1^{*\top}\mathcal{H}(\mathbf{y})_1\mathbf{x}_1^{*} \le  \mathbf{x}_1^{\top}\mathcal{H}(\mathbf{y})_1\mathbf{x}_1, \\
        0 &= \mathbf{x}_2^{*\top}\mathcal{H}(\mathbf{y})_2\mathbf{x}_2^{*} \le  \mathbf{x}_2^{\top}\mathcal{H}(\mathbf{y})_2\mathbf{x}_2, \\
        & \mathbf{x}^{\top}_1 \mathbf{A}_{3j1} \mathbf{x}_1 + \mathbf{x}^{\top}_2 \mathbf{A}_{3j2} \mathbf{x}_2 = b_{3j}.
    \end{aligned}
\end{equation}

This equation can also verify the stationary condition. Since $\mathbf{x}^{*}_1$ and $\mathbf{x}^{*}_2$ are the only nonzero solutions of $\mathbf{x}^{\top}\mathcal{H}(\mathbf{y})_1\mathbf{x} = 0$ and $\mathbf{x}^{\top}\mathcal{H}(\mathbf{y})_2\mathbf{x} = 0$ up to scale, respectively. The \ref{eq:Single Block SDP Problem} can always return the optimal solution with rank 1. $\hfill\square$

\section{Rounding and Nearest Rotation Recovery}
\label{supp:sec:Rounding and Nearest Rotation Recovery}
Due to observation noise and numerical approximation factors, the solutions obtained by the SDP optimizer may not strictly satisfy the rank-1 condition. Therefore, we need to employ a rounding strategy to recover from matrices approximated to rank 1. In addition, due to these approximate operations, the vanishing point matrix we obtain does not strictly adhere to the rotation matrix constraint. Hence, we need to employ a simple operation to restore the rotation matrix.

\subsection{Rounding Procedure}
Given a rank-1 approximation matrix $\mathbf{W} \in \mathbb{R}^{n\times n}$, the goal of the rounding operation is to obtain the optimal rank-1 approximation vector $\mathbf{w}$. We formulate the problem as follows:
\begin{equation}
    \begin{aligned}
        \mathbf{w} = arg \min_{\mathbf{w} \in \mathbb{R}^n} \quad & \left \| \mathbf{w}\mathbf{w}^{\top} - \mathbf{W} \right \|^{2}_{F}\\
         \text{s.t. } \quad &  \left \| \mathbf{w}\right \|^{2}_{2} = 1.
    \end{aligned}
\end{equation}
Utilizing the Frobenius norm, we can obtain an analytical solution that minimizes the norm error through Singular Value Decomposition (SVD) as:
\begin{equation}
    \begin{aligned}
        &\mathbf{U}\boldsymbol{\Sigma }\mathbf{V}^{\top} = \text{SVD}(\mathbf{W}),\\
        &\mathbf{w} = \sqrt{\sigma_1} [\mathbf{U}]_{*,1},
    \end{aligned}
\end{equation}
where $\sigma_1$ denotes the maximum singular value in $\boldsymbol{\Sigma}=\text{diag}(\sigma_1,\sigma_2,\sigma_3)$.

\subsection{Closed-Form Nearest Rotation Estimation}
Given a stacking rounding direction matrix $\mathbf{M} = [\mathbf{w}_1,\mathbf{w}_2,\mathbf{w}_3]$, the goal of the nearest rotation matrix estimation is to obtain a matrix $\mathbf{R} \in \mathbb{SO}(3) $ that satisfies the rotation matrix constraint with the minimum Frobenius norm error as:
\begin{equation}
    \begin{aligned}
        \mathbf{R} = arg \min_{\mathbf{R} \in \mathbb{SO}(3)} \quad & \left \| \mathbf{R} - \mathbf{M} \right \|^{2}_{F}
    \end{aligned}
\end{equation}
Similarly, the closed-form solution can be obtained as:
\begin{equation}
    \begin{aligned}
        &\mathbf{U}\boldsymbol{\Sigma }\mathbf{V}^{\top} = \text{SVD}(\mathbf{M}),\\
        &\mathbf{R} = \mathbf{U}\boldsymbol{\Sigma}^{'}\mathbf{V}^{\top},
    \end{aligned}
\end{equation}
where $\boldsymbol{\Sigma}^{'} = \text{diag}(1, 1, \text{det}(\mathbf{U}\mathbf{V}^{\top}))$.

\section{Additional Synthetic Experiments}
\label{supp:sec:Additional Synthetic Experiments}

\begin{figure*}[ht]
	\centering
	\includegraphics[width=0.99\linewidth]{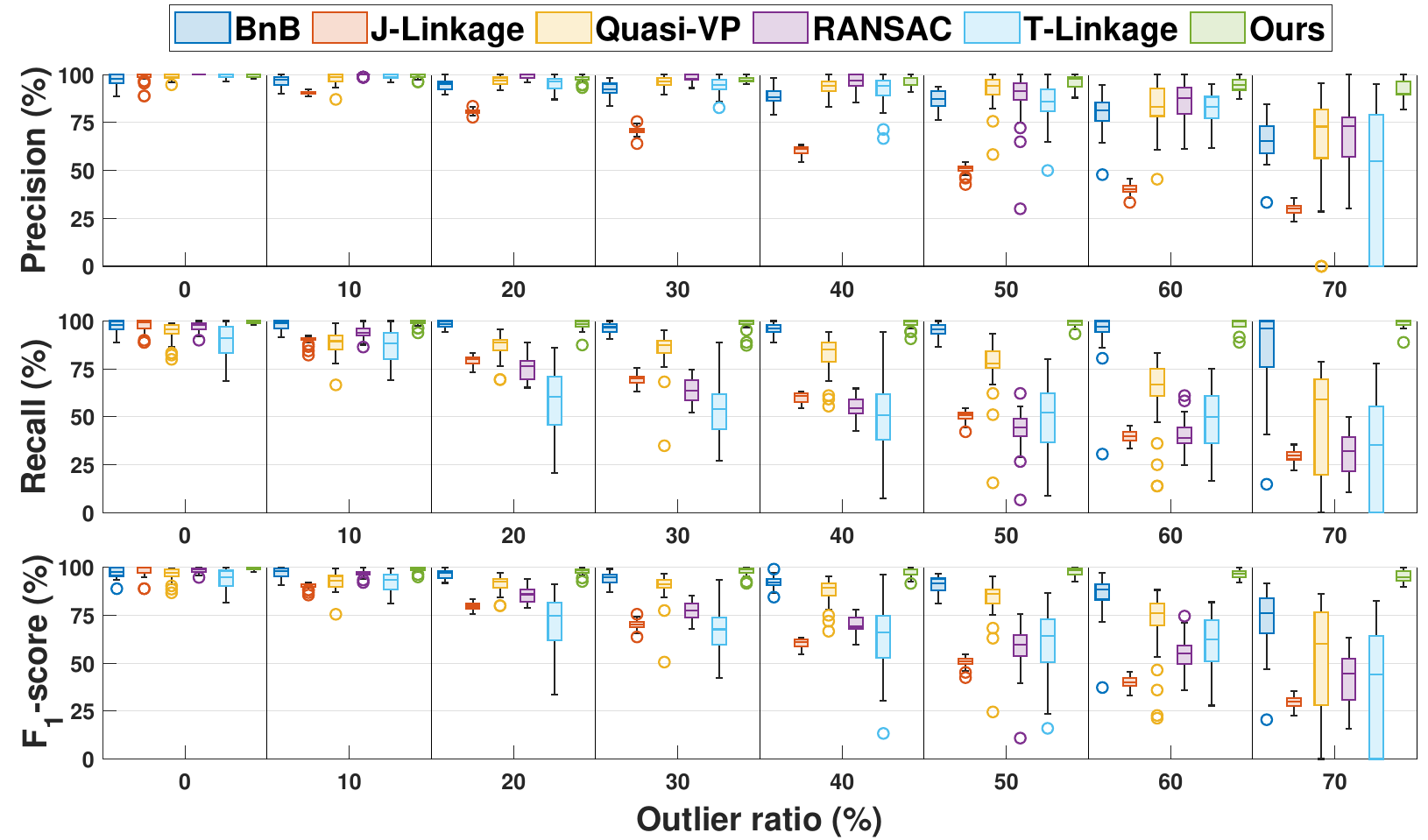}	\caption{Accuracy comparisons on the synthetic dataset with respect to the outlier ratios: boxplots of precision (top), recall (middle), and $F_1$-score (bottom). Best viewed in color and high resolution.}
	\label{fig_synthetic_complete_accuracy}
\end{figure*}

\PAR{Complete accuracy comparisons across outlier ratios.}
\label{supp:sec:Complete accuracy comparisons across outlier ratios}
We show the complete accuracy results across outlier ratios in \cref{fig_synthetic_complete_accuracy}. While only the $F_1$-score is shown in \cref{subsec:Synthetic Dataset}, this figure includes precision and recall as well. {\it \textbf{GlobustVP}} consistently demonstrates superior robustness and accuracy across varying outlier ratios.

\begin{figure*}[ht]
	\centering
	\includegraphics[width=0.99\linewidth]{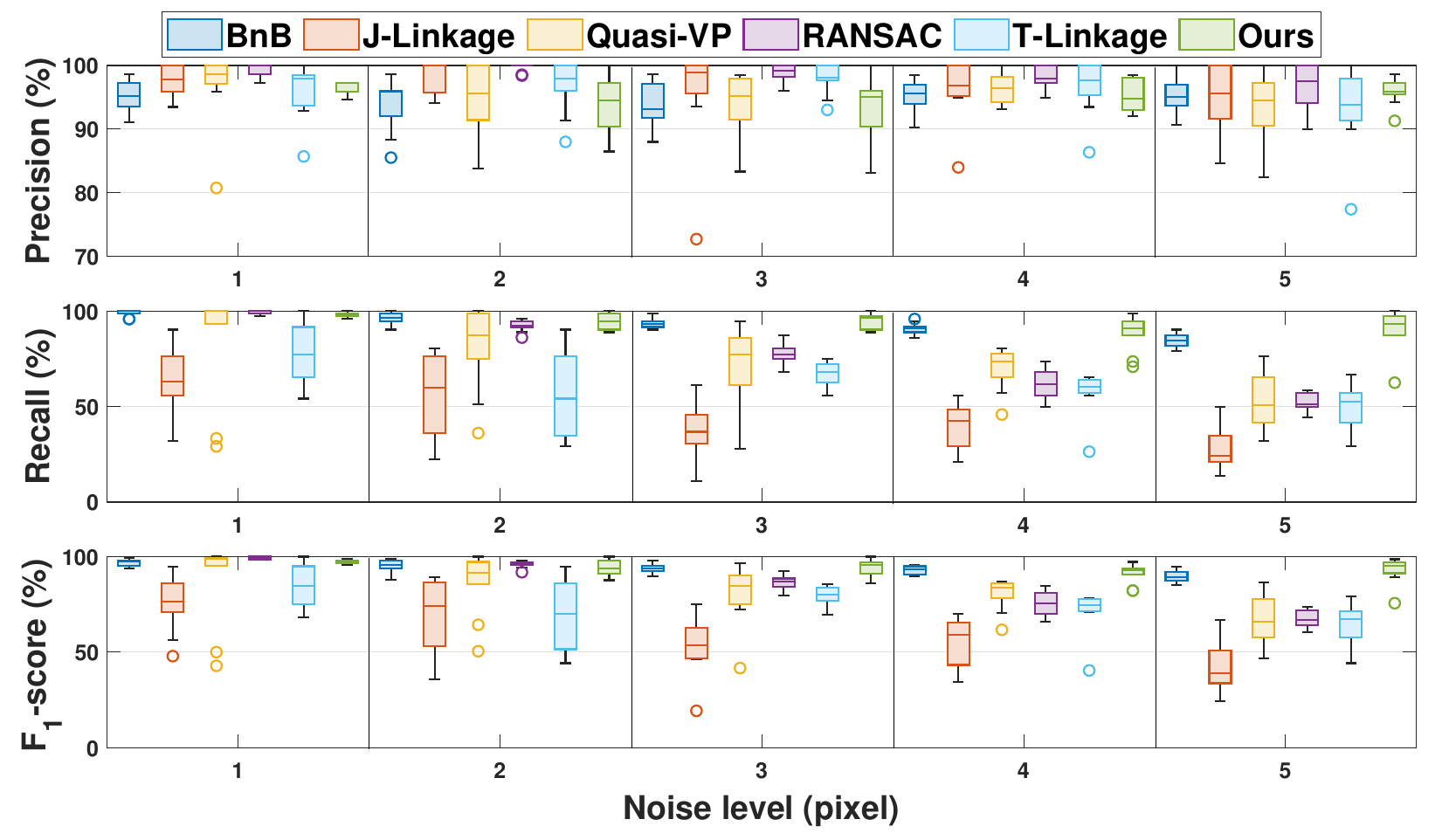}
	\caption{Accuracy comparisons on the synthetic dataset with respect to the noise levels: boxplots of precision (top), recall (middle), and $F_1$-score (bottom). Best viewed in color and high resolution.}
	\label{fig_synthetic_noise_acc}
\end{figure*}

\PAR{Accuracy comparisons across noise levels.}
We fix the number of lines at $60$ and the outlier ratio at $20\%$, while varying the standard deviation of zero-mean Gaussian noise from $1$ to $5$ pixels. \cref{fig_synthetic_noise_acc} shows the precision, recall, and $F_1$-score as the noise level increases. Recall and $F_1$-score reveal that \textbf{RANSAC}, \textbf{J-Linkage}, \textbf{T-Linkage}, and \textbf{Quasi-VP} are highly sensitive to noise. \textbf{BnB} begins to degrade significantly when the noise reaches $4$ pixel of noise. In contrast, {\it \textbf{GlobustVP}} maintains consistent performance across varying noise levels.

\begin{figure}[t]
	\centering
	\includegraphics[width=0.99\linewidth]{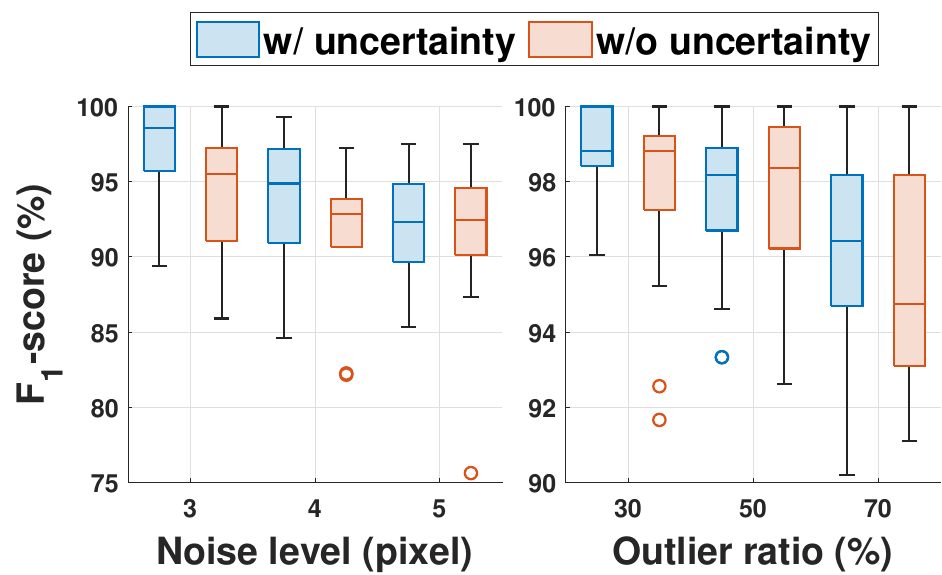}
	\caption{Comparisons of our method with (w/) and without (w/o) incorporating uncertainty. Best viewed in color.}
	\label{fig_uncertainty}
\end{figure}

\begin{figure*}[t]
	\centering
	\includegraphics[width=0.99\linewidth]{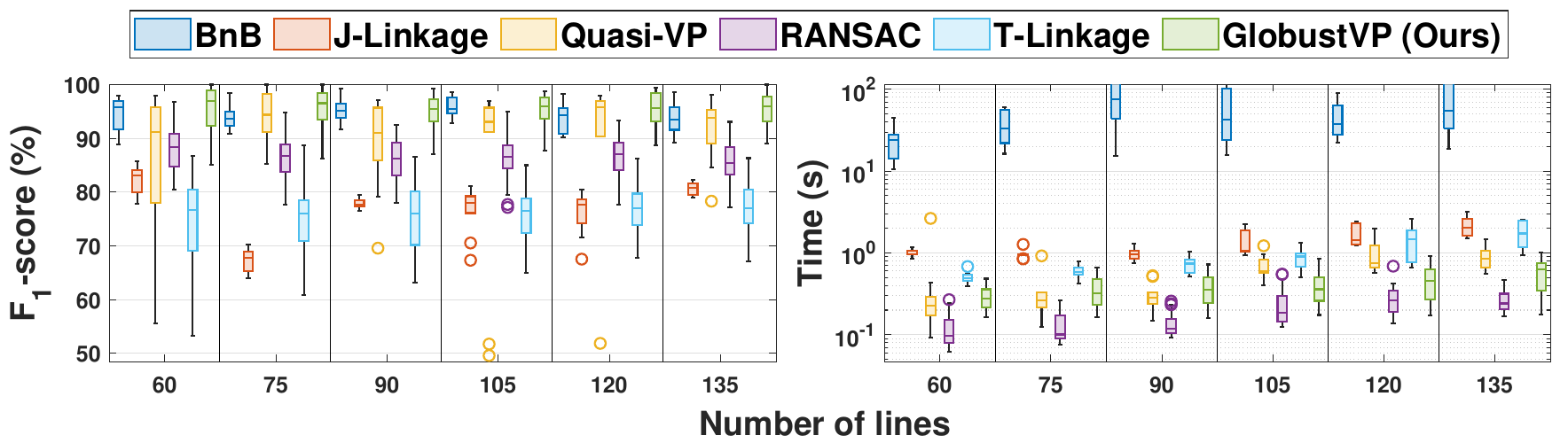}
	\caption{Accuracy and efficiency comparisons of all baseline methods on the synthetic dataset with respect to the number of image lines. Best viewed in color and high resolution.}
	\label{fig_synthetic_line_number}
\end{figure*}

\PAR{Complete accuracy and efficiency comparisons across line counts.}
In \cref{subsec:Synthetic Dataset}, we present comparisons of {\it \textbf{GlobustVP}} with \textbf{RANSAC} and \textbf{BnB} only. Here, we provide the complete accuracy and efficiency results for all methods. \cref{fig_synthetic_line_number} shows the $F_1$-score and runtime concerning the increasing number of lines with all methods. {\it \textbf{GlobustVP}} demonstrates consistent performance on both accuracy and efficiency as the number of lines increases.

\PAR{Ablation study on uncertainty.} Following~\cite{forstner2010optimal,forstner2016photogrammetric,heuel2001matching}, we derive the uncertainty for each line segment and incorporate it into $(\mathbf{d}^\top\mathbf{n})^2$.
We conduct an ablation study to evaluate its impact on accuracy (\ie, $F_1$-score) across 500 independent Monte Carlo trials with 60 lines. We vary the noise levels while keeping the outlier ratio fixed at $20\%$, and vary the outlier ratio while fixing the noise level at $\sigma = 3$. As shown in \cref{fig_uncertainty}, incorporating uncertainty slightly improves performance, while our original method (\ie, w/o uncertainty) achieves comparable results.

\section{Analysis of RANSAC-Based Methods}
\label{supp:sec:Analysis of RANSAC-based methods}

In \cref{subsec:Comparisons against Traditional Methods}, we use a RANSAC-based method~\cite{Zhang2016VanishingPE} (denoted as \textbf{RANSAC}) as a baseline for comparison with {\it \textbf{GlobustVP}}. The setup includes a maximum of 1,000 iterations, ensuring a 0.99 confidence level that a subset comprised solely of inliers is selected.

To ensure a fair and comprehensive comparison, we further analyze the performance of \textbf{RANSAC} and its variants. Specifically, we include two RANSAC variants: a slower version (denoted as \textbf{RANSAC-10K}), which allows up to 10,000 iterations to match the runtime of {\it \textbf{GlobustVP}}, and \textbf{MAGSAC}~\cite{barath2019magsac,barath2019magsacplusplus}, a state-of-the-art RANSAC-based method designed to improve accuracy.

We evaluate these methods on the YUD~\cite{Denis2008EfficientEM} and SU3~\cite{zhou2019learning} datasets, using the AA metric described in \cref{subsubsec:Real Experimental Settings}. As shown in \cref{tab_ransac_ablation}, \textbf{RANSAC} achieves the fastest runtime due to its limited iterations, but its accuracy is significantly lower compared to the other methods. \textbf{RANSAC-10K} improves accuracy over \textbf{RANSAC} on all metrics, albeit at the cost of increased runtime. On SU3~\cite{zhou2019learning}, \textbf{MAGSAC} surpasses \textbf{RANSAC-10K} in accuracy on all angular thresholds while consuming slightly more time. However, on YUD~\cite{Denis2008EfficientEM}, \textbf{RANSAC-10K} outperforms \textbf{MAGSAC} in AA$@3^{\circ}$ and AA$@10^{\circ}$.

Our proposed {\it \textbf{GlobustVP}} consistently outperforms all RANSAC-based methods on YUD~\cite{Denis2008EfficientEM} across all accuracy metrics. On SU3~\cite{zhou2019learning}, {\it \textbf{GlobustVP}} demonstrates superior performance in AA$@3^{\circ}$ and AA$@5^{\circ}$ while slightly lagging behind \textbf{MAGSAC} in AA$@10^{\circ}$. These results highlight the balanced trade-off between accuracy and runtime efficiency achieved by {\it \textbf{GlobustVP}} across both YUD~\cite{Denis2008EfficientEM} and SU3~\cite{zhou2019learning} datasets.

\begin{table}[h]
    \centering
    \setlength{\tabcolsep}{3pt}
    \resizebox{0.99\linewidth}{!}{
    \begin{tabular}{llcccc}
        \toprule
        Dataset & Method & AA$@3^{\circ}$ $\uparrow$ & AA$@5^{\circ}$ $\uparrow$ & AA$@10^{\circ}$ $\uparrow$ & Time (ms) $\downarrow$  \\
        \midrule
        \multirow{4}{*}{YUD~\cite{Denis2008EfficientEM}}
        & RANSAC~\cite{Zhang2016VanishingPE} & 53.3 & 64.7 & 78.4 & \cellcolor{tabfirst}9.0 \\
        & RANSAC-10K~\cite{Zhang2016VanishingPE} & \cellcolor{tabsecond}56.1 & 70.6 & \cellcolor{tabsecond}82.4 & 48.8 \\
        & MAGSAC~\cite{barath2019magsac} & 52.6 & \cellcolor{tabsecond}73.2 & 81.7 & \cellcolor{tabsecond}12.4 \\
        & GlobustVP (Ours) & \cellcolor{tabfirst}\textbf{67.6} & \cellcolor{tabfirst}\textbf{87.3} & \cellcolor{tabfirst}\textbf{96.1} & \textbf{48.8} \\
        \midrule
        \multirow{4}{*}{SU3~\cite{zhou2019learning}}
        & RANSAC [\textcolor{cyan}{10}] & 48.2 & 74.0 & 82.8 & \cellcolor{tabfirst}9.0 \\
        & RANSAC-10K [\textcolor{cyan}{10}] & 70.6 & 80.4 & 86.2 & 48.8 \\
        & MAGSAC [\textcolor{cyan}{1}] & \cellcolor{tabsecond}76.2 & \cellcolor{tabsecond}84.6 & \cellcolor{tabfirst}92.8 & \cellcolor{tabsecond}9.2 \\
        & GlobustVP (Ours) & \cellcolor{tabfirst}\textbf{80.2} & \cellcolor{tabfirst}\textbf{86.8} & \cellcolor{tabsecond}\textbf{92.4} & \textbf{48.8} \\
        \bottomrule
    \end{tabular}
    }
    \caption{Angular accuracy and runtime comparisons of RANSAC-based methods and {\it \textbf{GlobustVP}} on YUD~\cite{Denis2008EfficientEM} and SU3~\cite{zhou2019learning} datasets. The \colorbox{tabfirst}{best} and \colorbox{tabsecond}{second-best} performance for each metric are highlighted.}
    \label{tab_ransac_ablation}
\end{table}

\section{Additional Real-World Experiments}
\label{supp:sec:Additional Real-World Experiments}

\subsection{York Urban Database}
\label{supp:subsec:York Urban Database}

\begin{figure*}[t]
	\centering
	\includegraphics[width=0.99\linewidth]{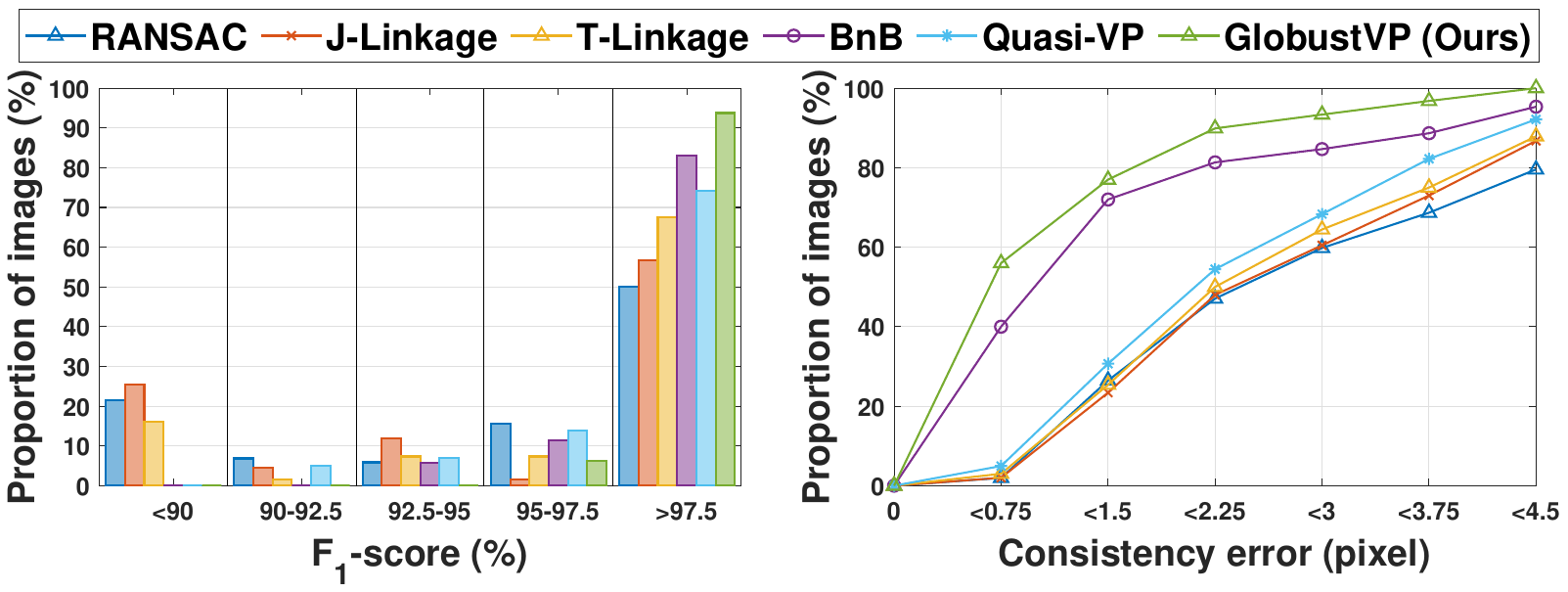}
	\caption{Accuracy comparisons on all images of YUD~\cite{Denis2008EfficientEM} using the manually extracted image lines. \textit{Left:} $F_1$-score of line-VP association. \textit{Right:} Cumulative histogram of the consistency error. Best viewed in color.}
	\label{fig_real_accuracy}
\end{figure*}

\begin{figure}[h]
	\centering
	\includegraphics[width=0.8\linewidth]{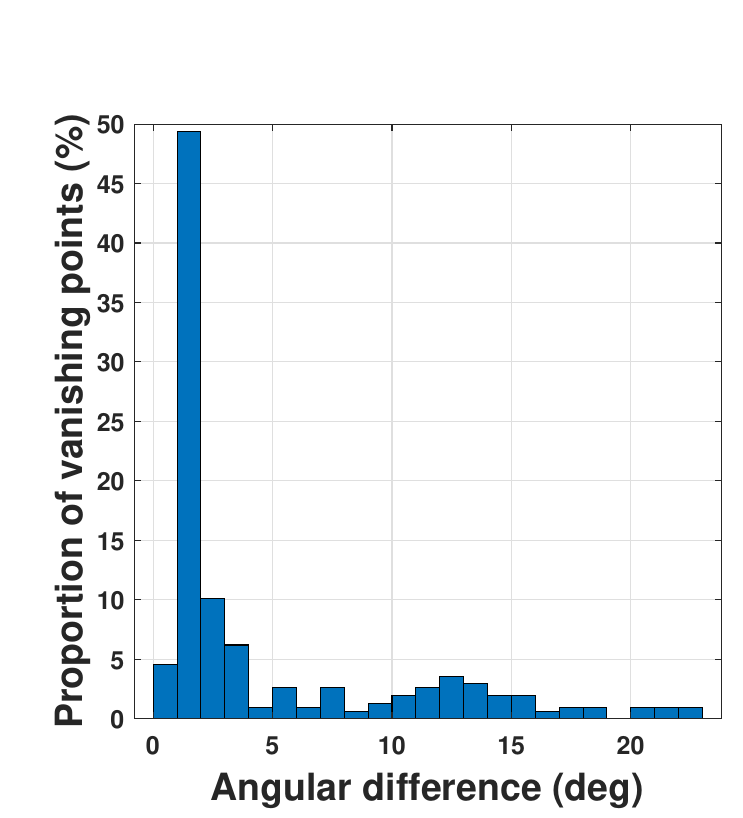}
	\caption{Histogram of angular differences between the estimated and the ground truth dominant directions on YUD~\cite{Denis2008EfficientEM}.}
	\label{fig_angular_histogram}
\end{figure}

\PAR{Accuracy comparisons.}
We provide a quantitative comparison of the baseline methods on YUD~\cite{Denis2008EfficientEM} in terms of $F_1$-score and consistency error, as shown in \cref{fig_real_accuracy}. We observe that \textbf{RANSAC}, \textbf{J-Linkge}, and \textbf{T-Linkage} struggle to achieve satisfactory accuracy, with up to $30\%$ images displaying a sub-optimal $F_1$-score (below $90\%$) and a large consistency error exceeding 2.25 pixels. While \textbf{BnB} demonstrates notable accuracy, it encounters convergence issues on a subset of images, leading to sub-optimal solutions. The performance of \textbf{Quasi-VP} falls between that of \textbf{RANSAC} and \textbf{BnB}, providing only a moderate level of accuracy. In contrast, {\it \textbf{GlobustVP}} achieves the highest accuracy by leveraging a global solver technique. Furthermore, to evaluate the accuracy of vanishing point estimation, we follow~\cite{forstner2010optimal} and compute the angular differences between the estimated and the ground truth dominant directions. The corresponding histogram of the angular differences on YUD~\cite{Denis2008EfficientEM} is shown in \cref{fig_angular_histogram}. The results demonstrate that approximately $50\%$ of the estimated dominant directions have angular difference less than $2^\circ$. This demonstrates that {\it \textbf{GlobustVP}} achieves excellent angular accuracy in vanishing point estimation.

\begin{figure*}[t]
	\centering
	\begin{subfigure}[b]{0.24\textwidth}
		\centering
        \caption*{\centering Lines}
		\includegraphics[width=\textwidth]{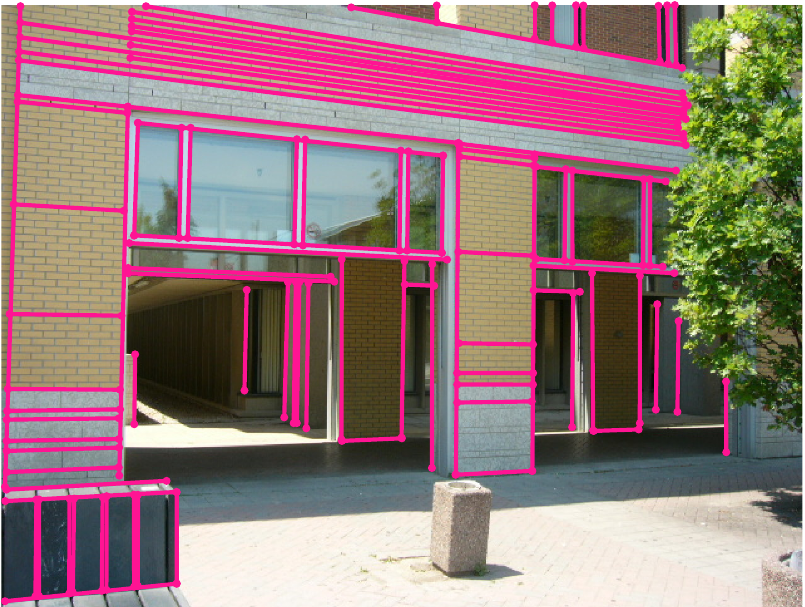}
		\caption*{\centering 2 VPs \par 95 lines}
		\label{fig_real_vis_manu_2vp_lines}
	\end{subfigure}
	\begin{subfigure}[b]{0.24\textwidth}
		\centering
		\caption*{\centering Ground Truth}
		\includegraphics[width=\textwidth]{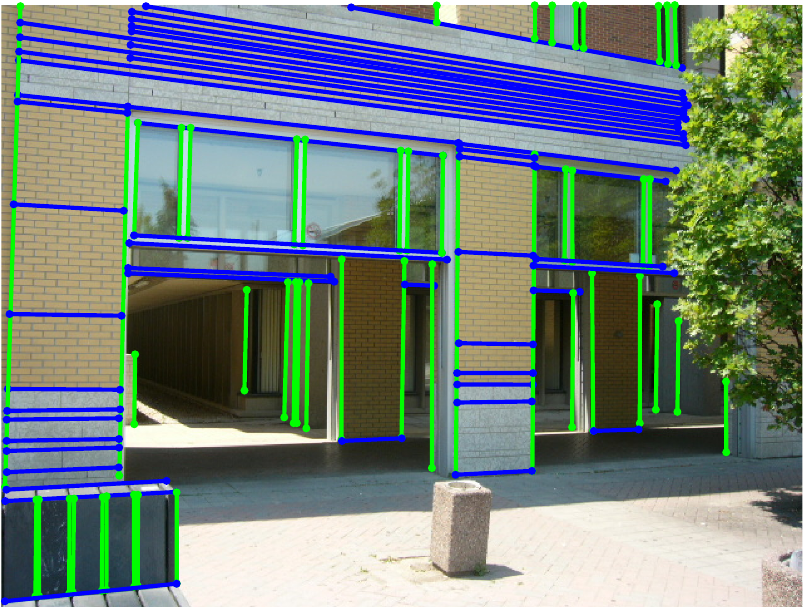}
		\caption*{\centering 2 VPs \par 95 lines}
		\label{fig_real_vis_manu_2vp_gt}
	\end{subfigure}
	\begin{subfigure}[b]{0.24\textwidth}
		\centering
		\caption*{\centering \textbf{RANSAC} \cite{Zhang2016VanishingPE}}
		\includegraphics[width=\textwidth]{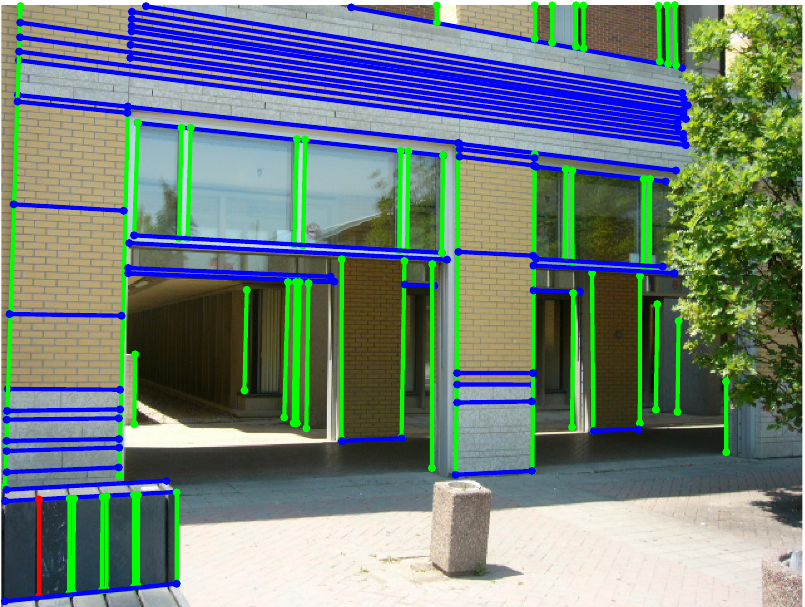}
        \centering
		\caption*{\centering \textcolor{pink}{$98.92\%$}, \textcolor{purple}{$96.84\%$} \par \textcolor{teal}{$97.87\%$}, \textcolor{violet}{0.43 pix}}
		\label{fig_real_vis_manu_2vp_ransac}
	\end{subfigure}
	\begin{subfigure}[b]{0.24\textwidth}
		\centering
		\caption*{\centering \textbf{J-Linkage} \cite{Toldo2008RobustMS}}
		\includegraphics[width=\textwidth]{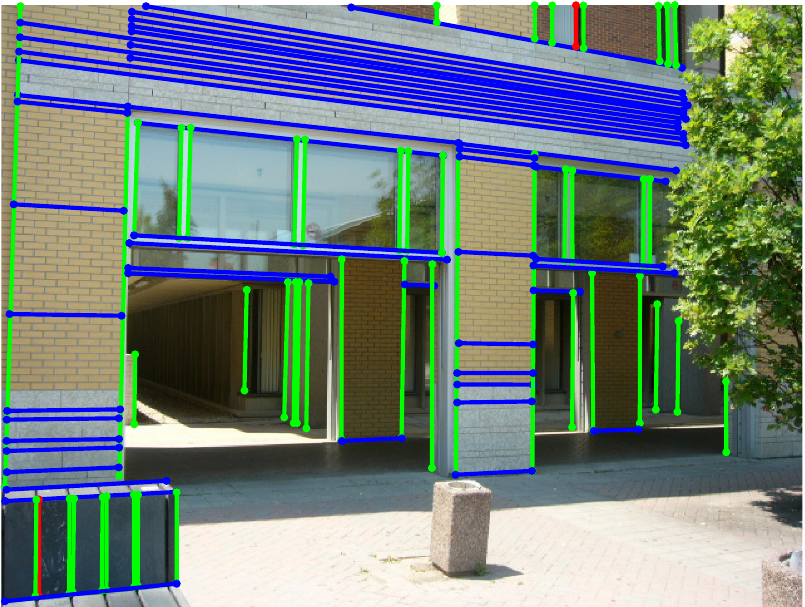}
		\caption*{\centering \textcolor{pink}{$97.87\%$}, \textcolor{purple}{$98.95\%$} \par \textcolor{teal}{$98.41\%$}, \textcolor{violet}{0.41 pix.}}
		\label{fig_real_vis_manu_2vp_j_linkage}
	\end{subfigure}
	
	\medskip
	
	\begin{subfigure}[b]{0.24\textwidth}
		\centering
		\caption*{\centering \textbf{T-Linkage} \cite{Magri2014TLinkageAC}}
		\includegraphics[width=\textwidth]{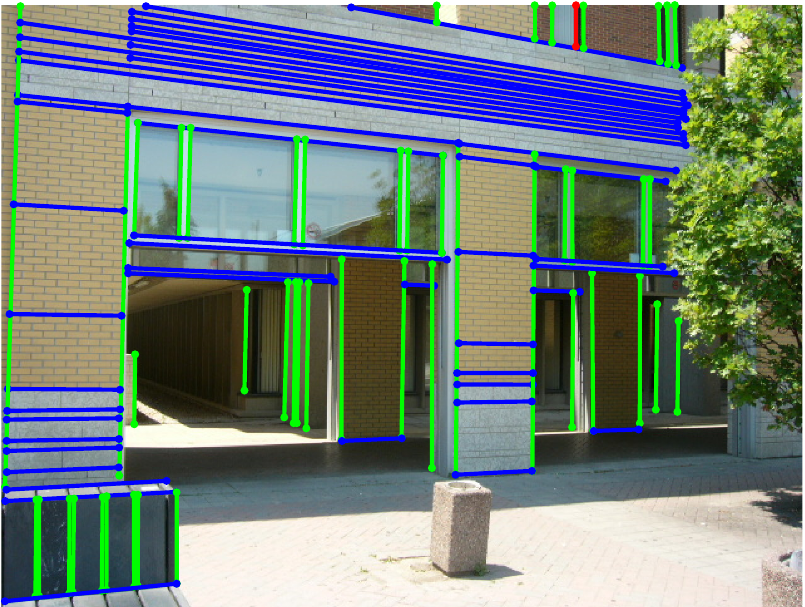}
		\caption*{\centering \textcolor{pink}{$98.92\%$}, \textcolor{purple}{$97.89\%$} \par \textcolor{teal}{$98.40\%$}, \textcolor{violet}{0.39 pix.}}
		\label{fig_real_vis_manu_2vp_t_linkage}
	\end{subfigure}
	\begin{subfigure}[b]{0.24\textwidth}
		\centering
		\caption*{\centering \textbf{BnB} \cite{Bazin2012GloballyOL}}
		\includegraphics[width=\textwidth]{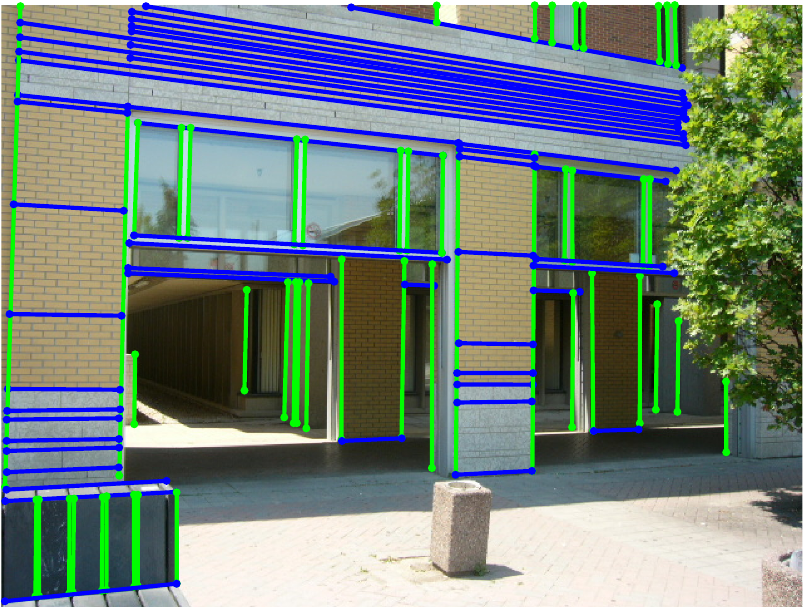}
		\caption*{\centering \textcolor{pink}{$100\%$}, \textcolor{purple}{$100\%$} \par \textcolor{teal}{$100\%$}, \textcolor{violet}{0.16 pix.}}
		\label{fig_real_vis_manu_2vp_bnb}
	\end{subfigure}
	\begin{subfigure}[b]{0.24\textwidth}
		\centering
		\caption*{\centering \textbf{Quasi-VP} \cite{Li2020QuasiGloballyOA}}
		\includegraphics[width=\textwidth]{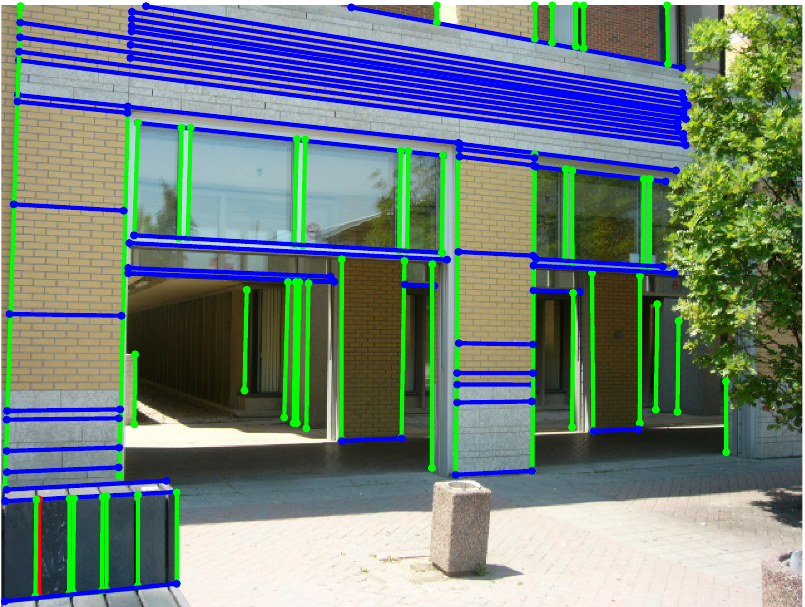}
		\caption*{\centering \textcolor{pink}{$97.22\%$}, \textcolor{purple}{$72.92\%$} \par \textcolor{teal}{$83.33\%$}, \textcolor{violet}{1.22 pix.}}
		\label{fig_real_vis_manu_2vp_quasi}
	\end{subfigure}
	\begin{subfigure}[b]{0.24\textwidth}
		\centering
		\caption*{\centering \textbf{GlobustVP (Ours)}}
		\includegraphics[width=\textwidth]{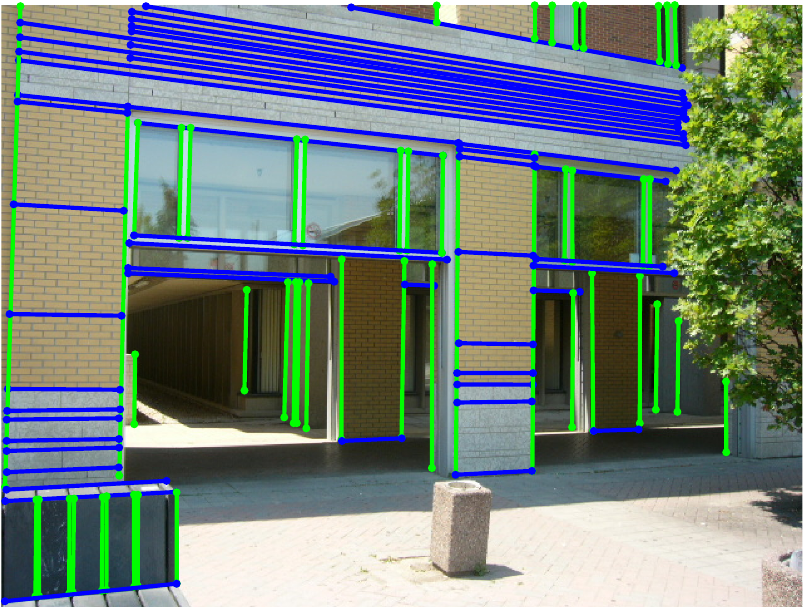}
		\caption*{\centering \textcolor{pink}{\textbf{$100\%$}}, \textcolor{purple}{\textbf{$100\%$}}, \par \textcolor{teal}{\textbf{$100\%$}}, \textcolor{violet}{\textbf{0.07 pix.}}}
		\label{fig_real_vis_manu_2vp_ours}
	\end{subfigure}
 
    \bigskip
 
    \begin{subfigure}[b]{0.24\textwidth}
		\centering
		\caption*{\centering Lines}
		\includegraphics[width=\textwidth]{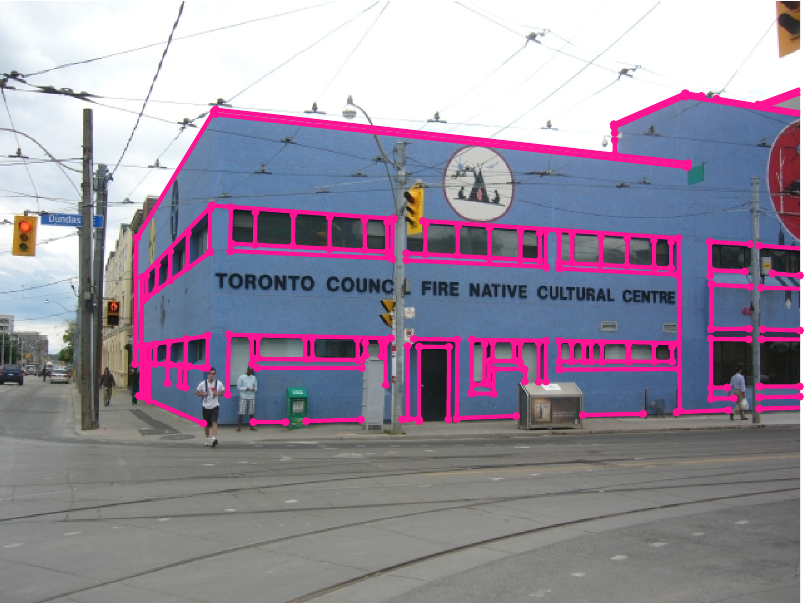}
		\caption*{\centering 3 VPs \par 129 lines}
		\label{fig_supp_real_vis_manu_3vp_lines}
	\end{subfigure}
	\begin{subfigure}[b]{0.24\textwidth}
		\centering
		\caption*{\centering Ground Truth}
		\includegraphics[width=\textwidth]{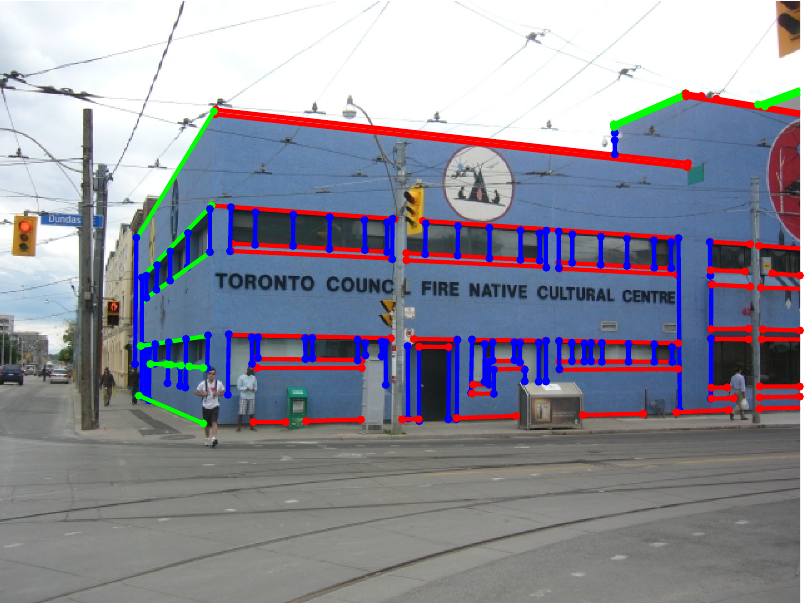}
		\caption*{\centering 3 VPs \par 129 lines}
		\label{fig_supp_real_vis_manu_3vp_gt}
	\end{subfigure}
	\begin{subfigure}[b]{0.24\textwidth}
		\centering
		\caption*{\centering \textbf{RANSAC} \cite{Zhang2016VanishingPE}}
		\includegraphics[width=\textwidth]{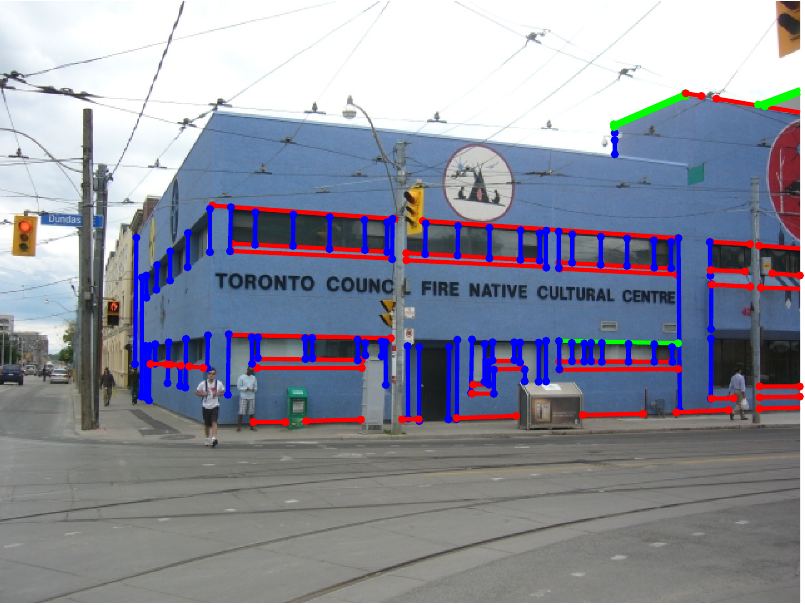}
		\centering
		\caption*{\centering \textcolor{pink}{$96.46\%$}, \textcolor{purple}{$84.50\%$} \par \textcolor{teal}{$90.08\%$}, \textcolor{violet}{0.45 pix.}}
		\label{fig_supp_real_vis_manu_3vp_ransac}
	\end{subfigure}
	\begin{subfigure}[b]{0.24\textwidth}
		\centering
		\caption*{\centering \textbf{J-Linkage} \cite{Toldo2008RobustMS}}
		\includegraphics[width=\textwidth]{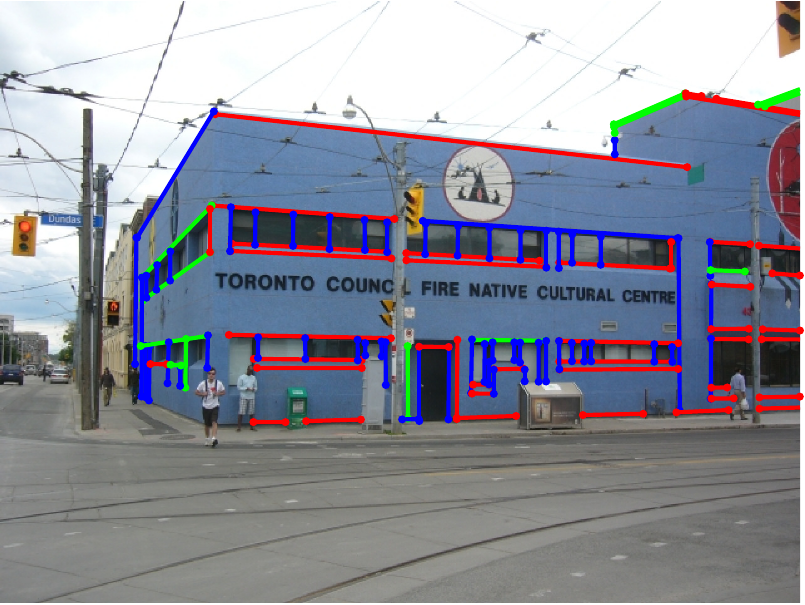}
		\caption*{\centering \textcolor{pink}{$89.09\%$}, \textcolor{purple}{$83.76\%$} \par \textcolor{teal}{$86.34\%$}, \textcolor{violet}{1.24 pix.}}
		\label{fig_supp_real_vis_manu_3vp_jlinkage}
	\end{subfigure}
	
	\medskip
	
	\begin{subfigure}[b]{0.24\textwidth}
		\centering
		\caption*{\centering \textbf{T-Linkage} \cite{Magri2014TLinkageAC}}
		\includegraphics[width=\textwidth]{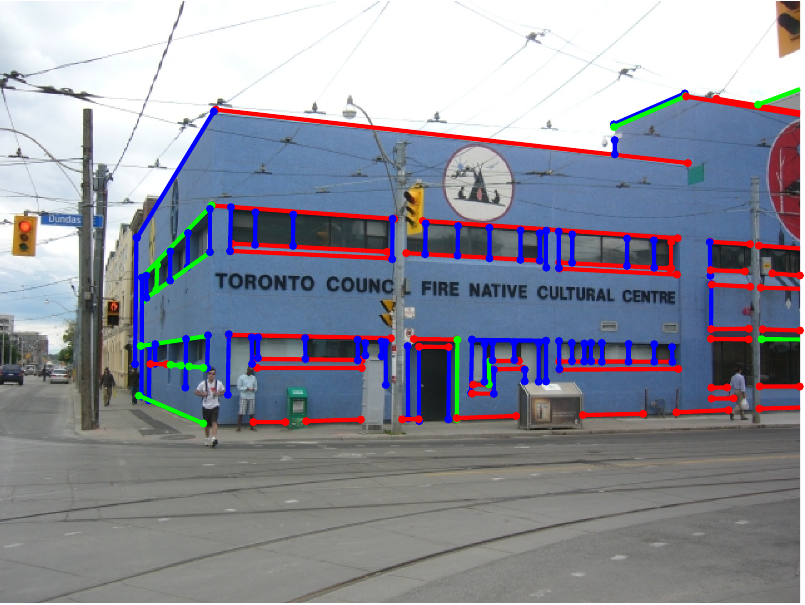}
		\caption*{\centering \textcolor{pink}{$91.89\%$}, \textcolor{purple}{$85.00\%$} \par \textcolor{teal}{$88.31\%$}, \textcolor{violet}{0.63 pix.}}
		\label{fig_supp_real_vis_manu_3vp_tlinkage}
	\end{subfigure}
	\begin{subfigure}[b]{0.24\textwidth}
		\centering
		\caption*{\centering \textbf{BnB} \cite{Bazin2012GloballyOL}}
		\includegraphics[width=\textwidth]{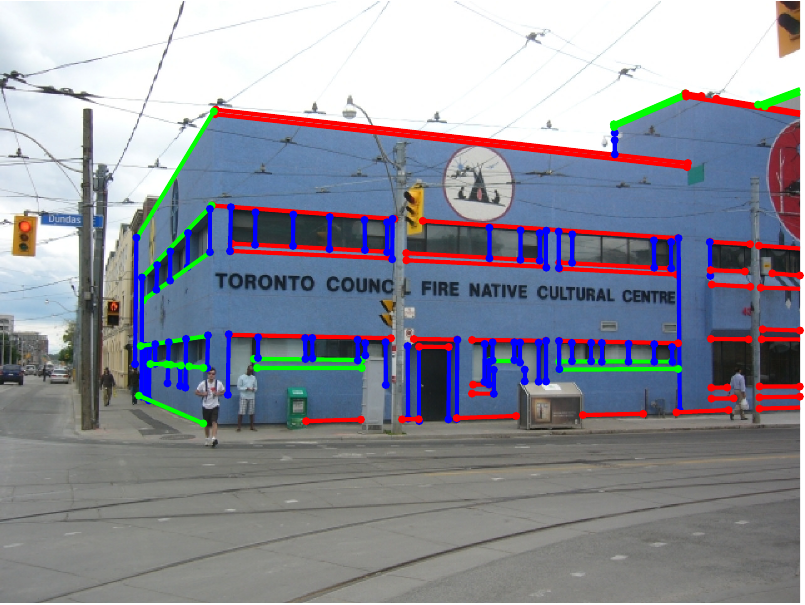}
		\caption*{\centering \textcolor{pink}{$87.60\%$}, \textcolor{purple}{$87.60\%$} \par \textcolor{teal}{$87.60\%$}, \textcolor{violet}{0.52 pix.}}
		\label{fig_supp_real_vis_manu_3vp_bnb}
	\end{subfigure}
	\begin{subfigure}[b]{0.24\textwidth}
		\centering
		\caption*{\centering \textbf{Quasi-VP} \cite{Li2020QuasiGloballyOA}}
		\includegraphics[width=\textwidth]{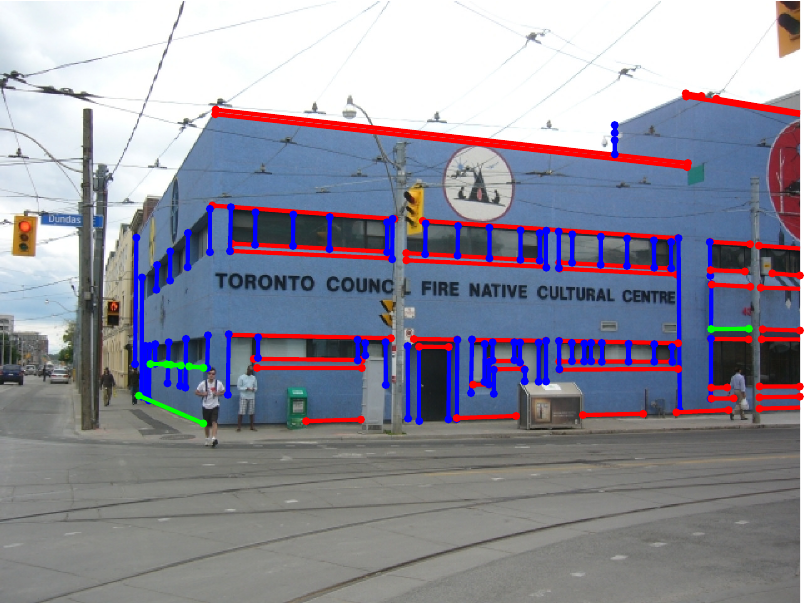}
		\caption*{\centering \textcolor{pink}{$99.11\%$}, \textcolor{purple}{$86.05\%$} \par \textcolor{teal}{$92.11\%$}, \textcolor{violet}{0.31 pix.}}
		\label{fig_supp_real_vis_manu_3vp_quasi}
	\end{subfigure}
	\begin{subfigure}[b]{0.24\textwidth}
		\centering
		\caption*{\centering \textbf{GlobustVP (Ours)}}
		\includegraphics[width=\textwidth]{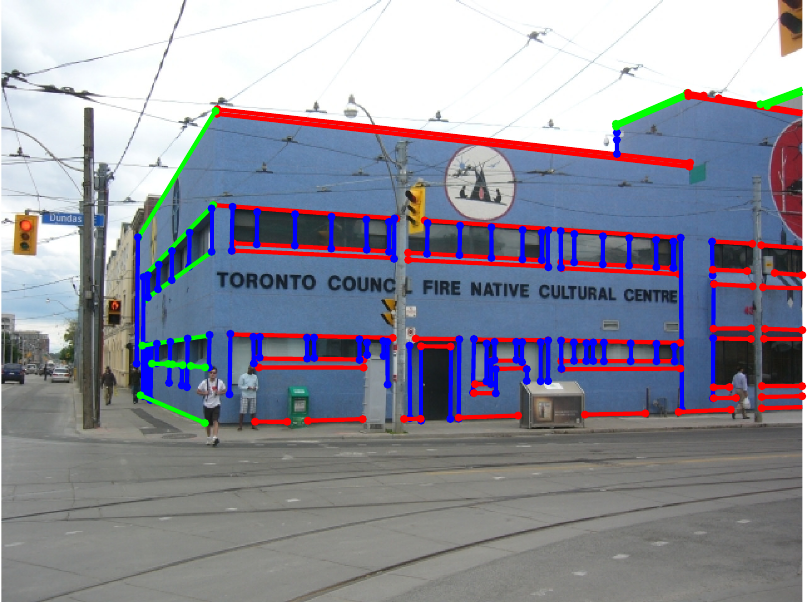}
		\caption*{\centering \textcolor{pink}{\textbf{$100\%$}}, \textcolor{purple}{\textbf{$100\%$}} \par \textcolor{teal}{\textbf{$100\%$}}, \textcolor{violet}{\textbf{0.01 pix.}}}
		\label{fig_supp_real_vis_manu_3vp_ours}
	\end{subfigure}
    \caption{Representative comparisons on YUD~\cite{Denis2008EfficientEM} using the manually extracted image lines. Different line-VP associations are shown in respective colors. The numbers below each image represent the respective \textcolor{pink}{precision $\uparrow$}, \textcolor{purple}{recall $\uparrow$}, \textcolor{teal}{$F_1$-score $\uparrow$}, and \textcolor{violet}{consistency error $\downarrow$} of line-VP association. Best viewed in color and high resolution.}
    \label{fig_supp_real_vis_manu}
\end{figure*}

\begin{figure*}[ht]
	\centering
	\begin{subfigure}[b]{0.18\textwidth}
		\centering
		\caption*{\centering Lines by LSD~\cite{Gioi2010LSDAF}}
		\includegraphics[width=\textwidth]{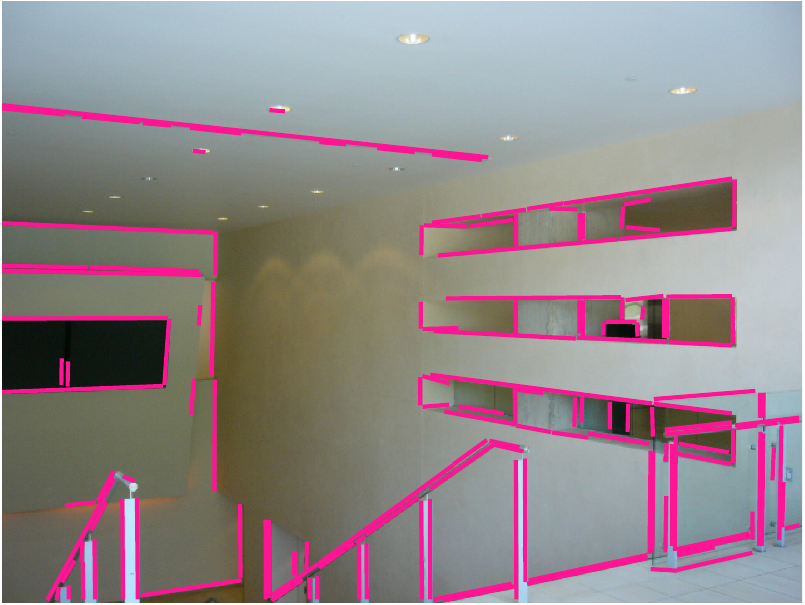}
		\caption*{\centering 3 VPs \par 148 lines}
		\label{fig_real_yud_vis_lsd_lines}
	\end{subfigure}
	\begin{subfigure}[b]{0.18\textwidth}
		\centering
		\caption*{\centering \textbf{RANSAC}~\cite{Zhang2016VanishingPE}}
		\includegraphics[width=\textwidth]{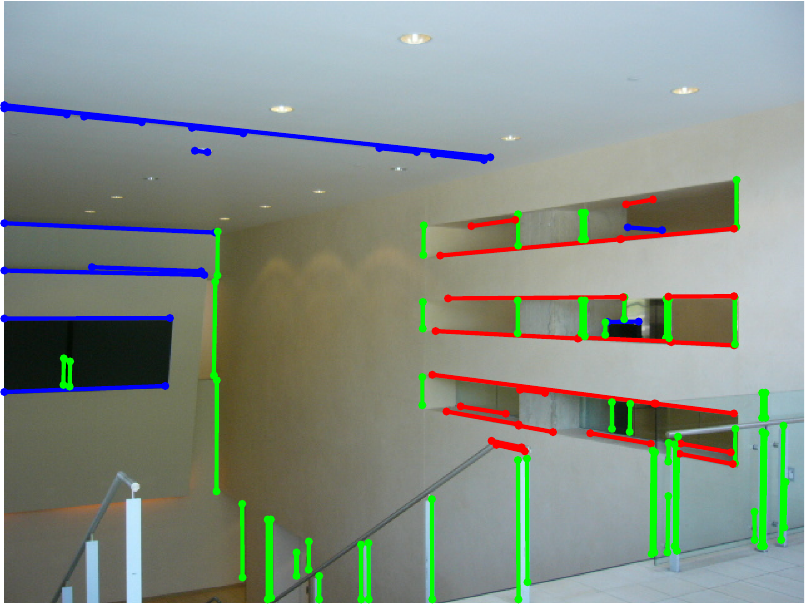}
		\centering
		\caption*{\centering \textcolor{pink}{$60.00\%$}, \textcolor{purple}{$86.17\%$} \par \textcolor{teal}{$70.74\%$}, \textcolor{violet}{0.27 pix.}}
		\label{fig_real_yud_vis_lsd_ransac}
	\end{subfigure}
	\begin{subfigure}[b]{0.18\textwidth}
		\centering
		\caption*{\centering \textbf{BnB}~\cite{Bazin2012GloballyOL}}
		\includegraphics[width=\textwidth]{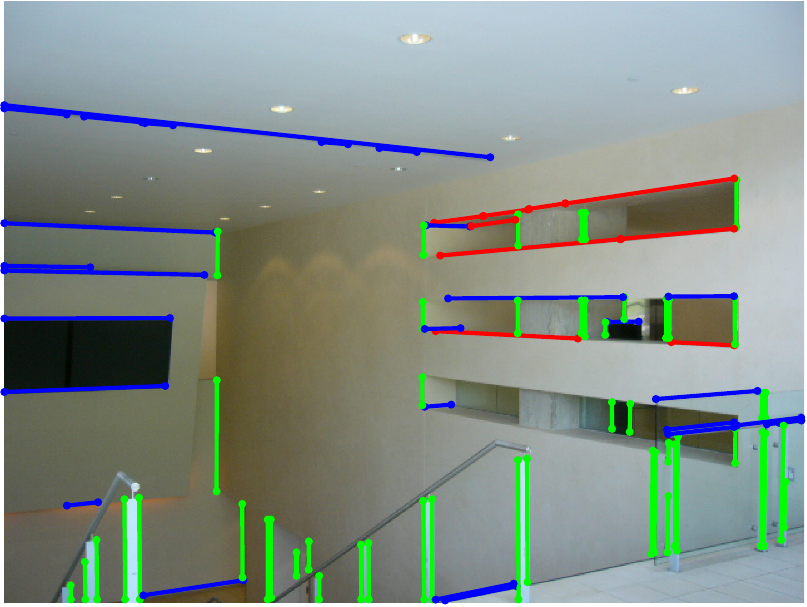}
		\caption*{\centering \textcolor{pink}{$69.11\%$}, \textcolor{purple}{$77.27\%$} \par \textcolor{teal}{$72.96\%$}, \textcolor{violet}{0.22 pix.}}
		\label{fig_real_yud_vis_lsd_bnb}
	\end{subfigure}
	\begin{subfigure}[b]{0.18\textwidth}
		\centering
		\caption*{\centering \textbf{Quasi-VP}~\cite{Li2020QuasiGloballyOA}}
		\includegraphics[width=\textwidth]{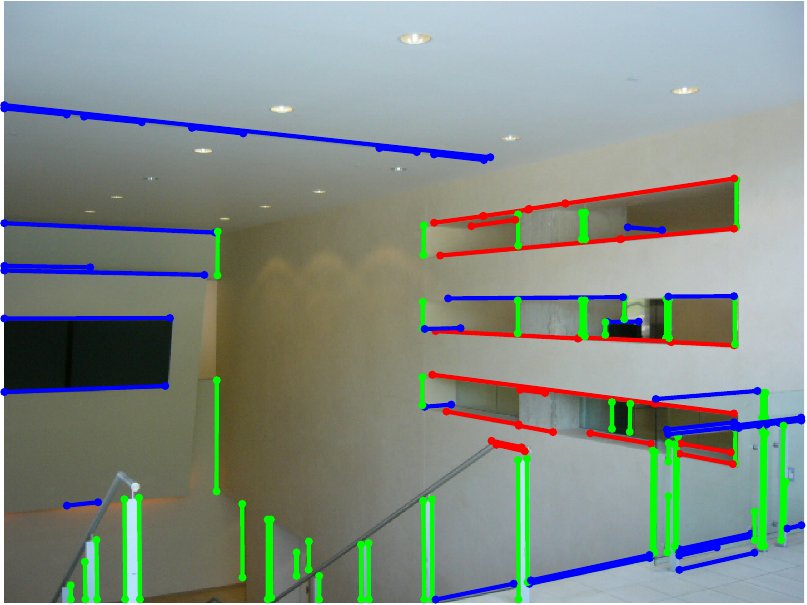}
		\caption*{\centering \textcolor{pink}{$100\%$}, \textcolor{purple}{$69.59\%$} \par \textcolor{teal}{$82.07\%$}, \textcolor{violet}{0.67 pix.}}
		\label{fig_real_yud_vis_lsd_quasi}
	\end{subfigure}
	\begin{subfigure}[b]{0.18\textwidth}
		\centering
		\caption*{\centering \textbf{GlobustVP (Ours)}}
		\includegraphics[width=\textwidth]{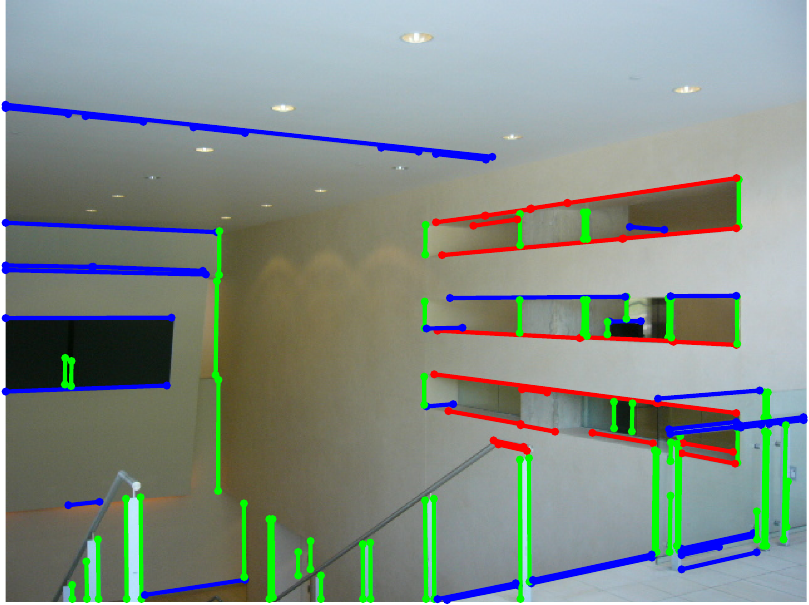}
		\caption*{\centering \textcolor{pink}{\textbf{$92.31\%$}}, \textcolor{purple}{\textbf{$77.70\%$}} \par \textcolor{teal}{\textbf{$84.38\%$}}, \textcolor{violet}{\textbf{0.13 pix.}}}
		\label{fig_real_yud_vis_lsd_ours}
	\end{subfigure}
	\caption{Representative comparisons on YUD~\cite{Denis2008EfficientEM} using the automatically extracted image lines by LSD~\cite{Gioi2010LSDAF}. Different line-VP associations are shown in respective colors. The numbers below each image represent the respective \textcolor{pink}{precision $\uparrow$}, \textcolor{purple}{recall $\uparrow$}, \textcolor{teal}{$F_1$-score $\uparrow$}, and \textcolor{violet}{consistency error $\downarrow$} of line-VP association. Best viewed in color and high resolution.}
	\label{fig_real_vis_lsd}
\end{figure*}

\PAR{Additional representative comparisons.}
We provide additional representative evaluation results on YUD~\cite{Denis2008EfficientEM} using the manually extracted image lines with $2$ and $3$ vanishing points (VPs), as shown in \cref{fig_supp_real_vis_manu}.
Furthermore, to further demonstrate the performance of {\it \textbf{GlobustVP}} on real-world images, we use YUD~\cite{Denis2008EfficientEM} and apply the Line Segment Detector (LSD)~\cite{Gioi2010LSDAF} to extract image lines. Subsequently, we estimate vanishing points using various methods and evaluate their accuracy. It is important to note that the extracted image lines inherently contain some outliers. The results in \cref{fig_real_vis_lsd} demonstrate a performance degradation of previous methods compared to those using the manually extracted image lines. Specifically, while \textbf{RANSAC} achieves a higher recall ($86.17\%$) due to its larger number of inliers, it also generates a greater number of incorrectly clustered lines, with a precision rate of $60.00\%$. Furthermore, \textbf{BnB} shows sub-optimal recall performance, as previously discussed in~\cite{Li2020QuasiGloballyOA}. In contrast, both \textbf{Quasi-VP} and {\it \textbf{GlobustVP}} achieve a balanced performance in terms of precision and recall. Notably, {\it \textbf{GlobustVP}} exhibits lower consistency error than other approaches, due to its inherent global optimality.

\subsection{NYU-VP Dataset}
\begin{figure*}[ht]
	\centering
	\begin{subfigure}[b]{0.18\textwidth}
		\centering
		\caption*{\centering Lines by LSD \cite{Gioi2010LSDAF}}
		\includegraphics[width=\textwidth]{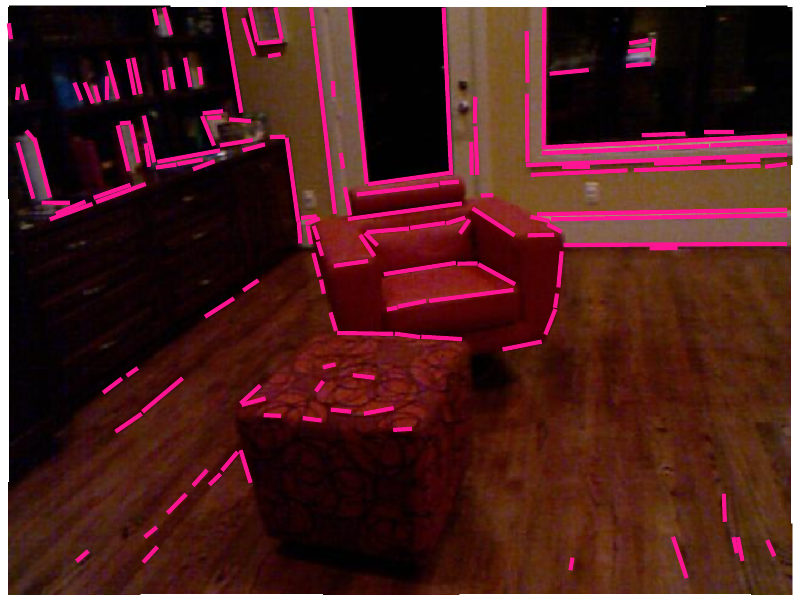}
		\caption*{\centering 3 VPs, 150 lines}
		\label{fig_real_nyu_vis_lsd_lines_1}
	\end{subfigure}
	\begin{subfigure}[b]{0.18\textwidth}
		\centering
		\caption*{\centering \textbf{RANSAC} \cite{Zhang2016VanishingPE}}
		\includegraphics[width=\textwidth]{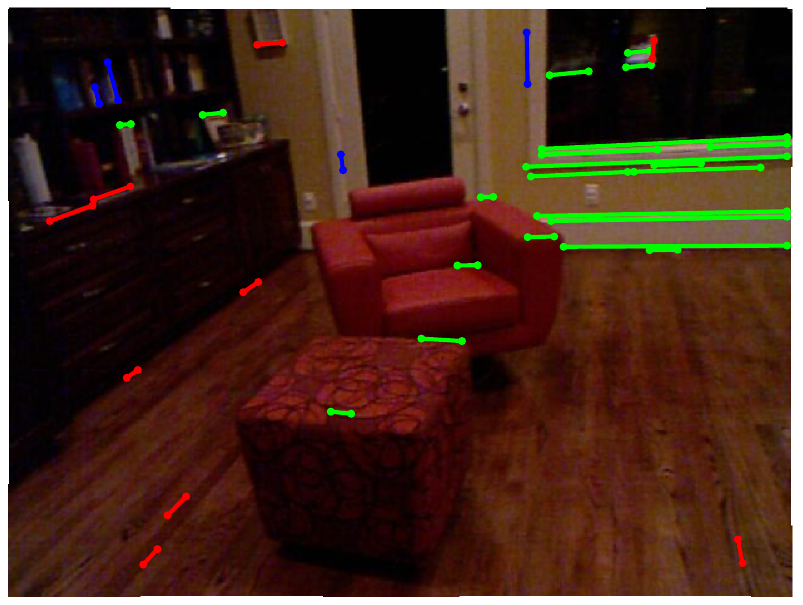}
		\centering
		\caption*{\centering $34$ inliers}
		\label{fig_real_nyu_vis_lsd_ransac_1}
	\end{subfigure}
	\begin{subfigure}[b]{0.18\textwidth}
		\centering
		\caption*{\centering \textbf{BnB} \cite{Bazin2012GloballyOL}}
		\includegraphics[width=\textwidth]{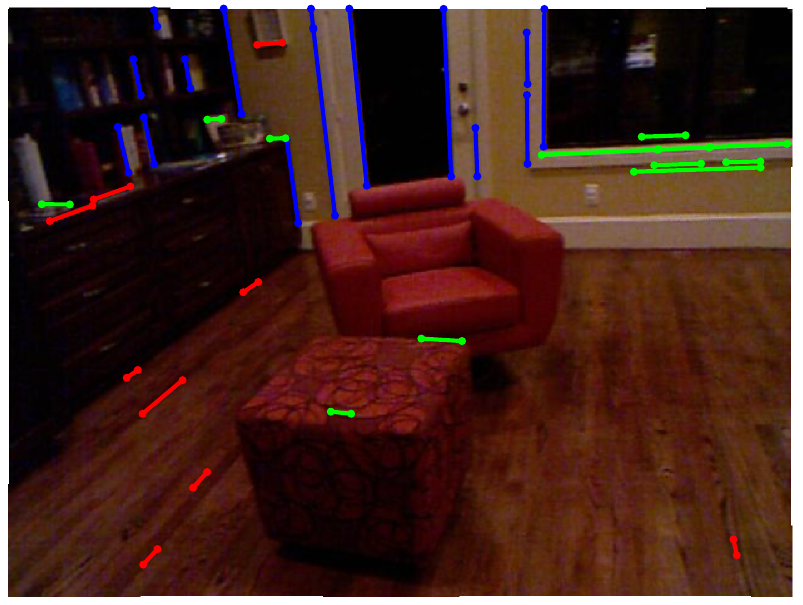}
		\caption*{\centering $36$ inliers}
		\label{fig_real_nyu_vis_lsd_bnb_1}
	\end{subfigure}
	\begin{subfigure}[b]{0.18\textwidth}
		\centering
		\caption*{\centering \textbf{Quasi-VP} \cite{Li2020QuasiGloballyOA}}
		\includegraphics[width=\textwidth]{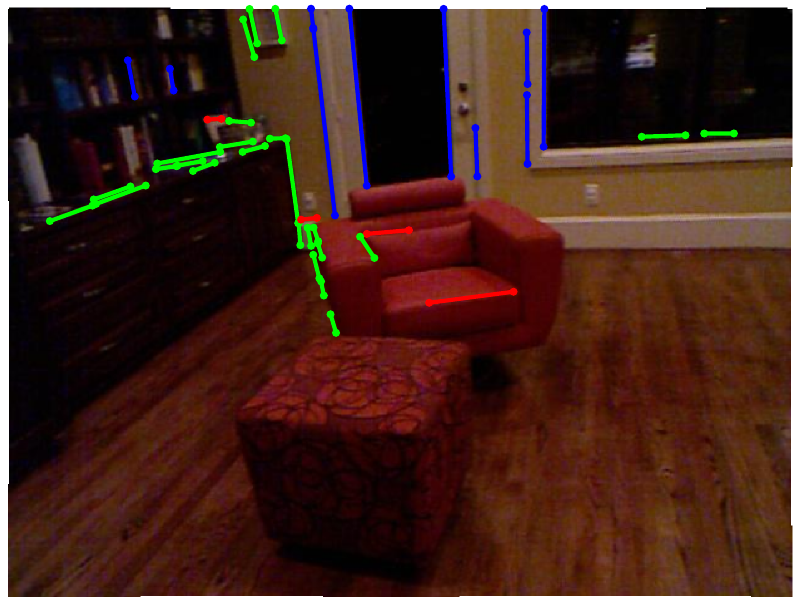}
		\caption*{\centering $39$ inliers}
		\label{fig_real_nyu_vis_lsd_quasi_1}
	\end{subfigure}
	\begin{subfigure}[b]{0.18\textwidth}
		\centering
		\caption*{\centering \textbf{GlobustVP (Ours)}}
		\includegraphics[width=\textwidth]{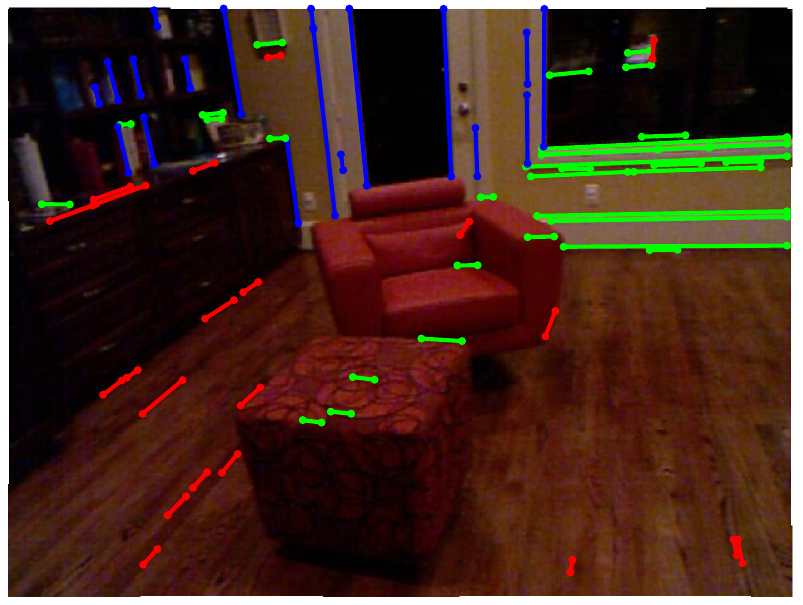}
		\caption*{\centering \textbf{$\textbf{70}$ inliers}}
		\label{fig_real_nyu_vis_lsd_ours_1}
	\end{subfigure}

    \medskip

    \begin{subfigure}[b]{0.18\textwidth}
		\centering
		\includegraphics[width=\textwidth]{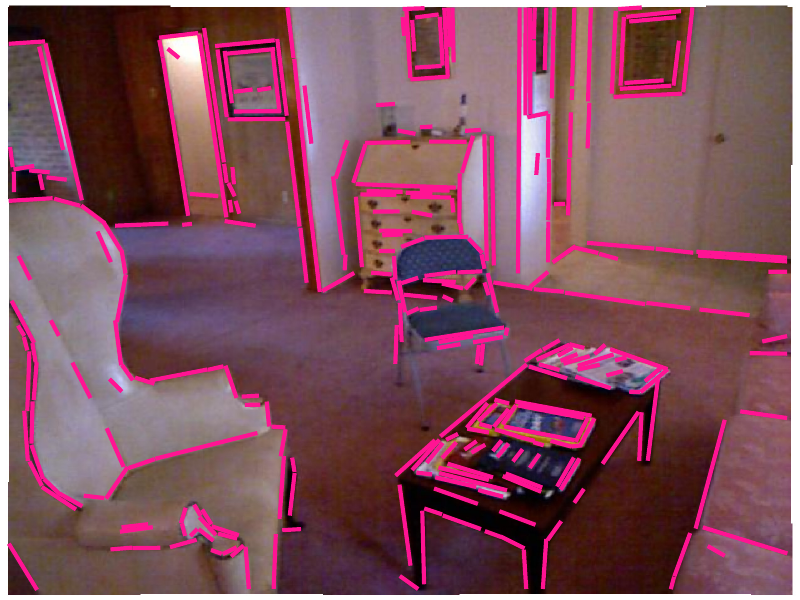}
		\caption*{\centering 3 VPs, 304 lines}
		\label{fig_real_nyu_vis_lsd_lines_2}
	\end{subfigure}
	\begin{subfigure}[b]{0.18\textwidth}
		\centering
		\includegraphics[width=\textwidth]{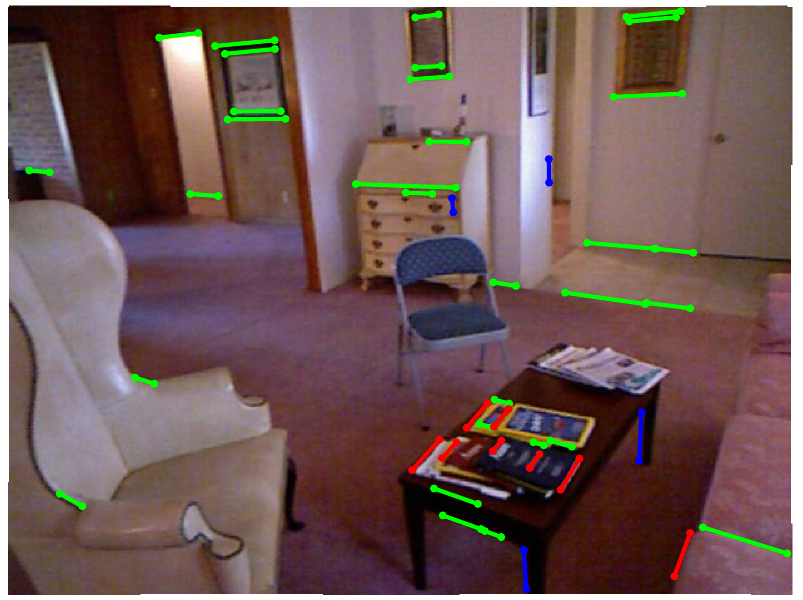}
		\centering
		\caption*{\centering $43$ inliers}
		\label{fig_real_vis_nyu_lsd_ransac_2}
	\end{subfigure}
	\begin{subfigure}[b]{0.18\textwidth}
		\centering
		\includegraphics[width=\textwidth]{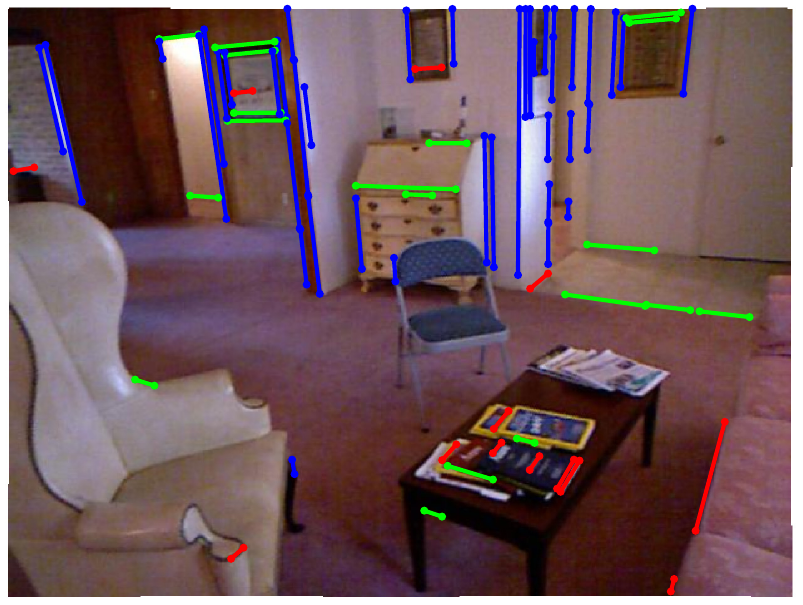}
		\caption*{\centering $72$ inliers}
		\label{fig_real_vis_nyu_lsd_bnb}
	\end{subfigure}
	\begin{subfigure}[b]{0.18\textwidth}
		\centering
		\includegraphics[width=\textwidth]{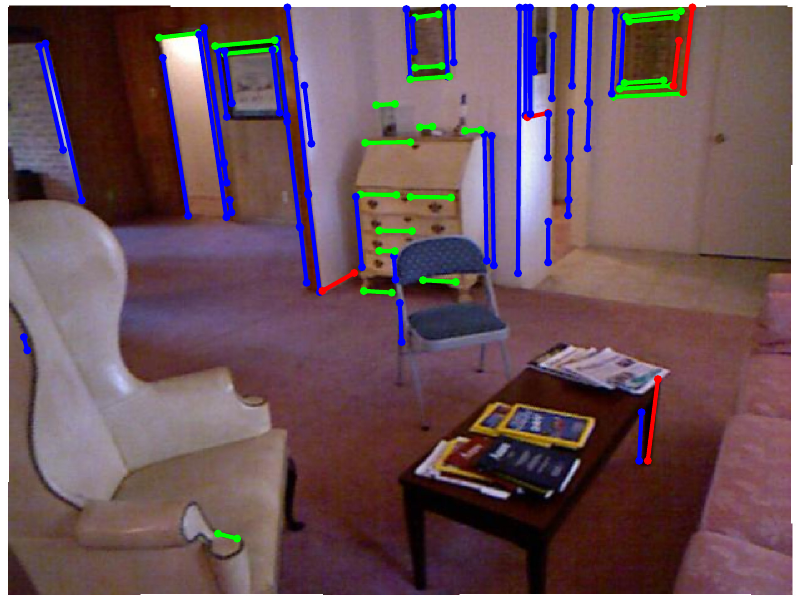}
		\caption*{\centering $70$ inliers}
		\label{fig_real_vis_nyu_lsd_quasi_2}
	\end{subfigure}
	\begin{subfigure}[b]{0.18\textwidth}
		\centering
		\includegraphics[width=\textwidth]{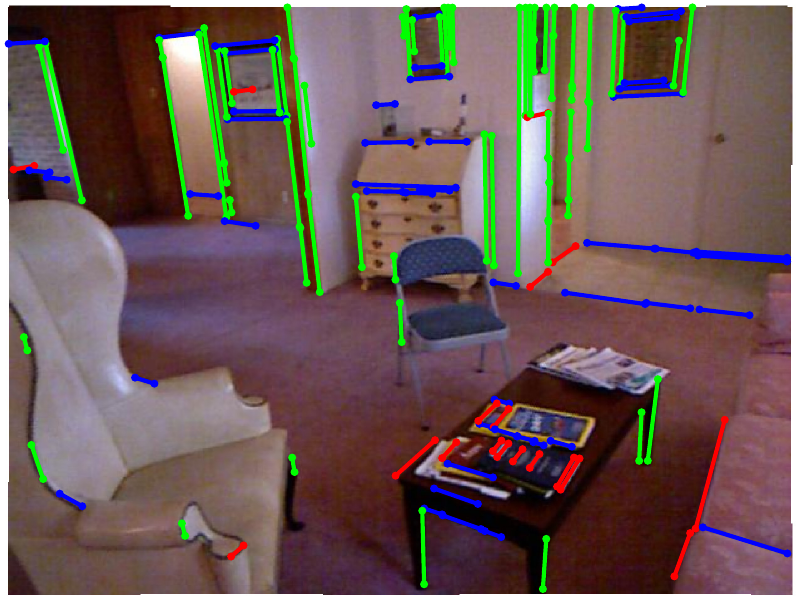}
		\caption*{\centering \textbf{$\textbf{127}$ inliers}}
		\label{fig_real_vis_nyu_lsd_ours_2}
	\end{subfigure}
	\caption{Representative comparisons on NYU-VP dataset~\cite{Kluger2020CONSACRM} using the automatically extracted image lines by LSD \cite{Gioi2010LSDAF}. Different line-VP associations are shown in respective colors. The number below each image represents the number of inliers identified by each method. Best viewed in color and high resolution.}
	\label{fig_real_nyu_vis_lsd}
\end{figure*}

To further evaluate the performance of {\it \textbf{GlobustVP}}, we conduct additional experiments on the NYU-VP dataset~\cite{Kluger2020CONSACRM}. The dataset contains ground truth vanishing points for 1449 indoor scenes, with line segments extracted from the images using LSD~\cite{Gioi2010LSDAF}. The results shown in \cref{fig_real_nyu_vis_lsd} demonstrate that {\it \textbf{GlobustVP}} produces a higher number of inliers than previous approaches.

\clearpage\clearpage
{
    \small
    \bibliographystyle{ieeenat_fullname}
    \bibliography{main}
}

\end{document}